\documentclass{article}

\usepackage{dsfont}  
\usepackage{booktabs}  
\usepackage{amsmath}  
\usepackage{graphicx}  
\usepackage{subcaption} 
\usepackage[ruled,vlined, noend]{algorithm2e}
\usepackage{wrapfig}  
\usepackage[preprint]{corl_2021} 
\usepackage[symbol]{footmisc}
\usepackage[title]{appendix}

\title{Learning Eye-in-Hand Camera Calibration\\from a Single Image}

\author{
  Eugene Valassakis\footnotemark[1]\\
  The Robot Learning Lab \\
  Imperial College London \\
  United Kingdom\\
  \texttt{pev115@ic.ac.uk} \\
  \And
  Kamil Drezckowski\footnotemark[1]\\
  The Robot Learning Lab \\
  Imperial College London \\
  United Kingdom\\
  \texttt{krd115@ic.ac.uk} \\
  \And
    Edward Johns\\
  The Robot Learning Lab \\
  Imperial College London \\
  United Kingdom\\
  \texttt{e.johns@imperial.ac.uk} \\
}

\begin{document}

%
%
\maketitle
\vspace{-0.5cm}
\begin{abstract} Eye-in-hand camera calibration is a fundamental and long-studied problem in robotics. We present a study on using learning-based methods for solving this problem online from a single RGB image, whilst training our models with entirely synthetic data. We study three main approaches: one direct regression model that directly predicts the extrinsic matrix from an image, one sparse correspondence model that regresses 2D keypoints and then uses PnP, and one dense correspondence model that uses regressed depth and segmentation maps to enable ICP pose estimation. In our experiments, we benchmark these methods against each other and against well-established classical methods, to find the surprising result that direct regression outperforms other approaches, and we perform noise-sensitivity analysis to gain further insights into these results.

\end{abstract}


%
%
\section{Introduction}

\vspace{-0.3cm}
\footnotetext[1]{Joint First Author Contribution}

\begin{wrapfigure}{r}{0.5\linewidth}
\vspace{-0.9cm}
\centering
\includegraphics[width = \linewidth]{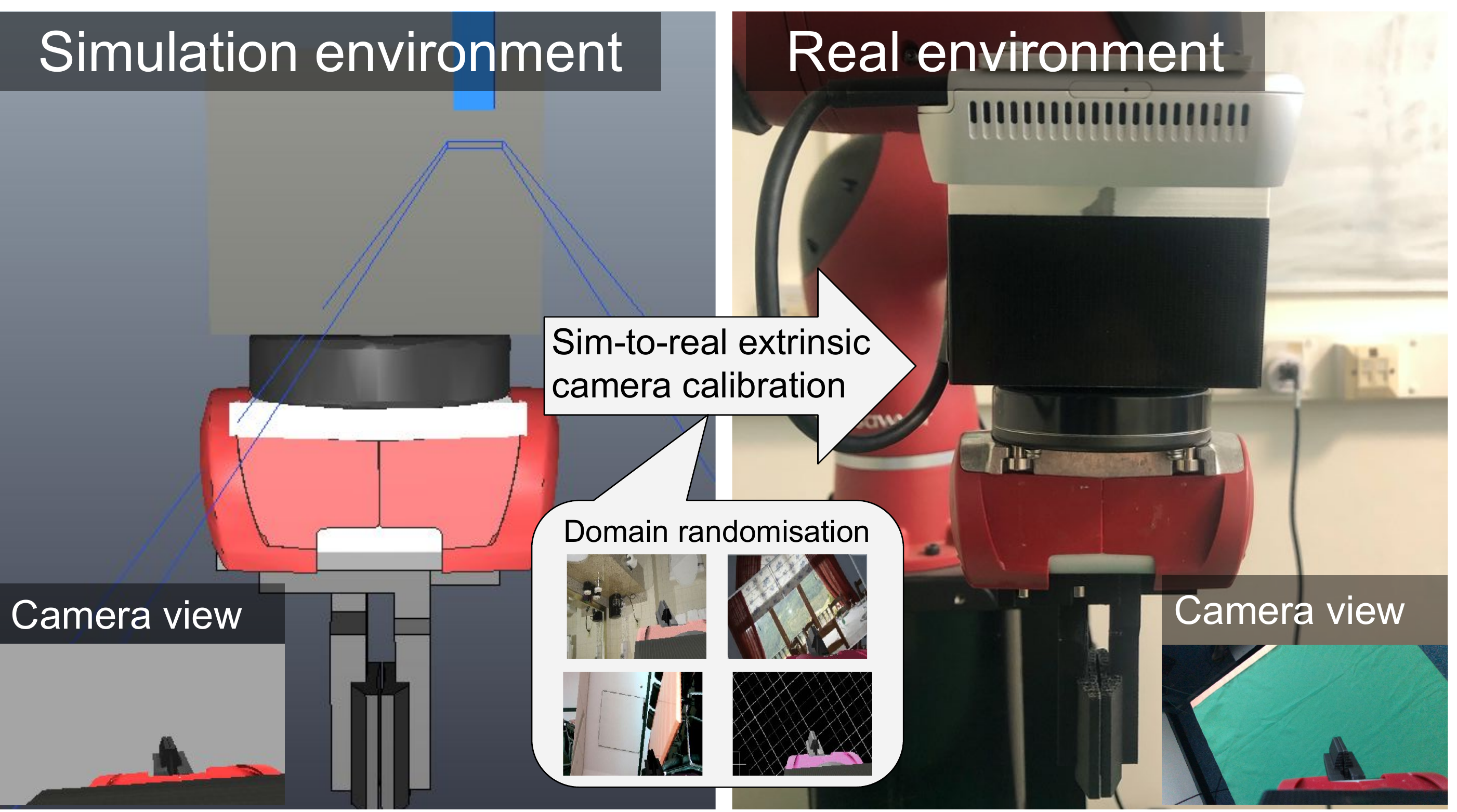}
\caption{Illustration of our task and setup.}
\vspace{-0.3cm}
\label{fig:dislay_fig}
\end{wrapfigure}


Eye-in-hand setups, where images from a camera rigidly mounted to the end-effector are used to control a robot, are a popular choice for precise robot manipulation. Compared to control using a fixed external camera, the eye-in-hand setup enables precise control because images can be captured very close to an object, as well as generalisation across the workspace when actions are defined relative to the camera or end-effector~\citep{johns2021coarse, cam-cali-for-manipulation, johns2016deep}. To apply to precise manipulation tasks, these methods require accurate eye-in-hand extrinsic calibration, to ensure that actions or poses inferred in the camera's frame can be transformed to the robot's frame. Whilst calibration-free approaches exist, such as end-to-end control~\cite{valassakis2021coarse, puang2020kovis}, these are only suitable for model-free policy learning and cannot be used for methods that involve explicit planning or reasoning about 3D space.

Classically, solving the eye-in-hand camera calibration problem is performed off-line~\cite{daniilidis1999hand,tsai1989new,park1994robot,horaud1995hand} and involves the use of a specific external calibration object, such as a checkerboard or an AprilTag. Poses of the calibration object can be estimated by the camera, and by recording them from several robot configurations, the rigid-body transformation between the camera and end-effector can be estimated analytically. Although with perfect information these methods can yield exact calibrations, they are in practice limited by the quality of the data they use, including how precisely the pose of the calibration object can be estimated or the accuracy of the recorded end-effector poses. Furthermore, these methods are unsuited for use by non-experts since they require dedicated external calibration objects and a new calibration dataset to be carefully collected for each re-calibration. Finally, they are impractical as re-calibration needs to be performed every time the camera position changes due to (1) re-mounting, (2) wear and tear, and (3) collisions between the camera and the environment. Deep learning methods can alleviate these drawbacks. Given the initial overhead required to generate training data and to train a model, a deep learning-based method can be readily used to re-calibrate a camera online from a single image using parts of the robot as the calibration object. This potential has been demonstrated for \textit{eye-to-hand} camera calibration, where an external camera observes the robot from a distance~\cite{labbe2021single,lee2020camera}.

To the best of our knowledge, there are currently no studies on using deep learning methods for \textit{eye-in-hand} calibration, which is the focus of this paper. We propose that this calibration can be estimated directly from the camera's image itself, without requiring any external apparatus, as long as the image can observe part of the end-effector, such as the gripper's fingers (see fig.~\ref{fig:dislay_fig}). Based on this, we identify and benchmark three natural approaches for leveraging the power of deep learning in this way: two based on classical, geometric pose estimation where deep learning is used in place of components that are typically manually engineered, and one which uses deep learning to directly regress the camera calibration matrix. We show in our real-world experiments that the two approaches that rely on geometric analysis perform poorly, and we analyse empirically the structural limitations and challenges that cause this. We also show that, perhaps surprisingly, direct regression outperforms all alternatives, including classical methods based on fiducial markers with automatic data collection. Finally, given that direct regression can estimate the calibration matrix from a single image online, we propose and evaluate a method for fusing multiple estimates of the camera's pose to increase the accuracy of the aggregated estimate. 

As such, our work has three key contributions: (1) we investigate the suitability of three natural alternatives for using deep-learning for enabling eye-in-hand camera calibration from a single image, (2) we show the potential of end-to-end deep learning for online eye-in-hand camera calibration in everyday environments, and (3) we provide an analysis for the two learning-based methods that rely on geometric approaches, through which we discover why these methods may not be suitable for eye-in-hand camera calibration. An accompanying video can be found at  \newline \url{https://www.robot-learning.uk/learning-eye-in-hand-calibration}~.
%
%
\section{Related Work}\label{sec:related_work}

%
%
Eye-in-hand camera calibration is a long-studied problem, with significant advances introduced in the 1990's, with what are now well-established solutions. The problem consists of inferring the camera to end-effector pose for a camera mounted on the wrist of a robotic manipulator. To do so, several end-effector to robot base poses are recorded, along with corresponding estimates of calibration object to camera poses. The problem is then reduced to solving the $AX=XB$ equation through a formalism introduced by Shiu and Ahmad in \citep{shiu1987calibration}. Several works then followed that mainly differ in their strategy for solving the $AX=XB$ equation. For instance, Tsai and Lenz~\citep{tsai1989new} improve on the efficiency of Shiu and Ahmad's solution by proposing a closed-form solution. Park and Martin~\citep{park1994robot} and Dornaika and Harod~\citep{horaud1995hand} also consider such a closed-form solution while relying on Lie theory and unit quaternions respectively. 

While the above approaches solve for rotation and translation separately, Dornaika and Harod~\citep{horaud1995hand}  also propose a non-linear technique for solving for both simultaneously. Daniilidis~\citep{daniilidis1999hand} proposes a solution to the problem using dual quaternions.  Finally, more recent works study various extensions for particular settings such as using structure for motion for calibration~\cite{andreff2001robot}, using model-based pose estimation and tracking for online calibration~\cite{pauwels2016integrated}, simultaneously considering the data time synchronisation problem~\cite{furrer2018evaluation}, calibrating depth sensors with non-overlapping views~\cite{faion2012recursive}, automatically doing both calibration pattern localisation and eye-in-hand camera calibration~\cite{antonello2017fully}, and optimising the robot kinematics parameters and the camera to end-effector pose simultaneously~\cite{liu2020fast}.

Recently, deep learning methods have had great successes on closely related tasks such as pose estimation~\citep{li2018deepim,labbe2020cosypose} and end-to-end robotics manipulation from wrist-mounted cameras~\citep{,johns2021coarse, valassakis2021coarse}. Moreover, sim-to-real transfer has shown promise in alleviating the large data requirements that hinder the scalability of such methods~\citep{valassakis2021coarse, valassakis2020crossing,tobin2017domain, alghonaim2021benchmarking}. In this work, we investigate the naturally emerging question of how one might use such methods for eye-in-hand camera calibration, which is fundamental to a large number of robotics pipelines. Closest to our work, Lee et al.~\citep{lee2020camera} and Labbe et al.~\citep{labbe2021single} investigate two different deep learning approaches for eye-to-hand calibration of a camera looking straight at a robot. \citep{lee2020camera} is based on keypoint regression followed by Perspective-n-Point (PnP) \citep{PnP} while \citep{labbe2021single} adopts a render-and-compare approach. In our work, we consider the eye-in-hand setup that is more common when precise manipulation needs to be achieved by looking closely at the end-effector of the robot~\citep{valassakis2021coarse, haugaard2020fast}. This problem also differs in its properties from the eye-to-hand setup, since in a typical image there is now a much more constrained view that only allows for the tip of the end-effector to be visible, and used as an anchor for pose estimation. 
%
%
\section{Methods}\label{sec:method}
%
%
\begin{figure}
    \centering
    \includegraphics[width=\linewidth]{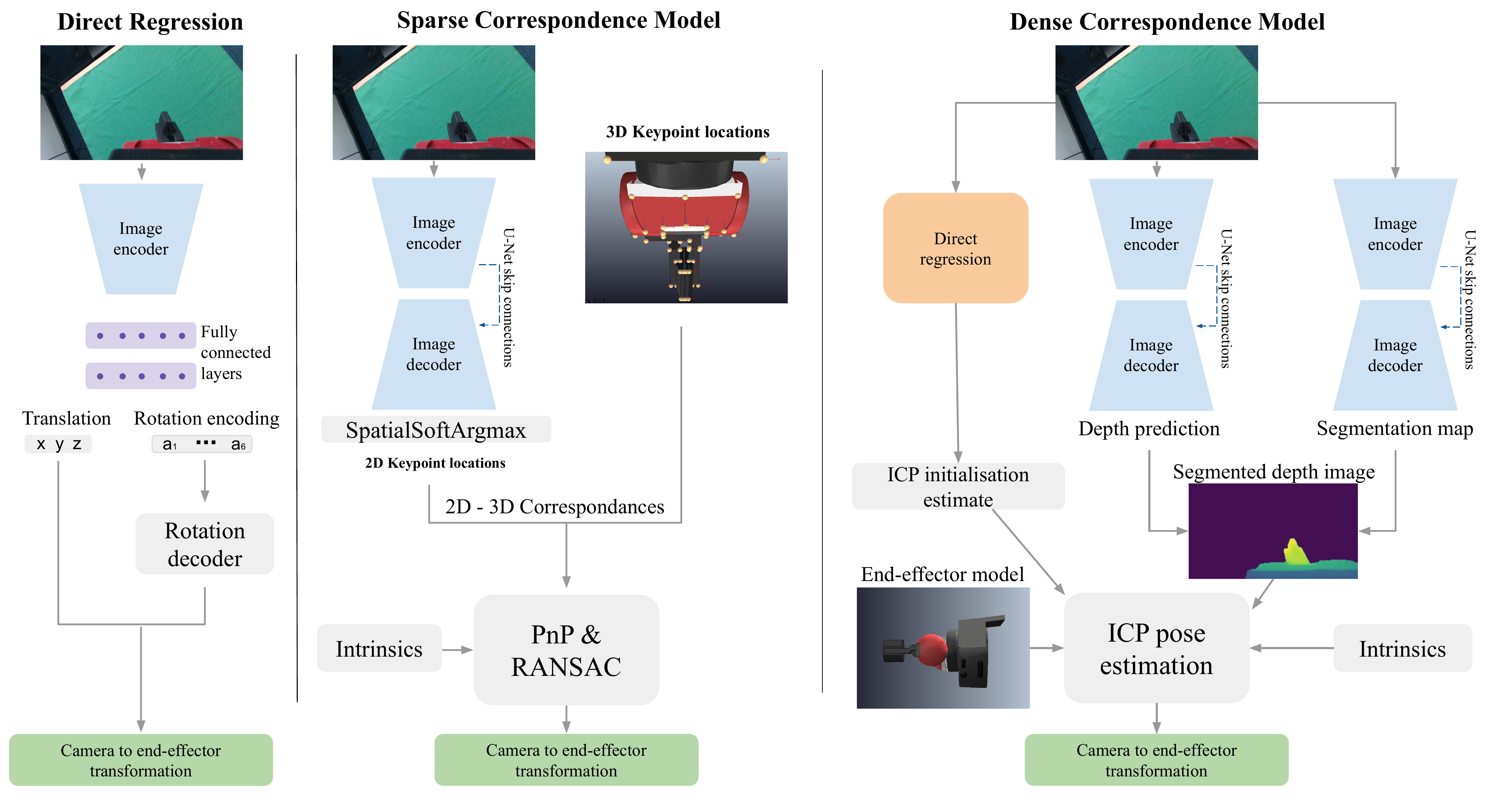}
    \caption{From left to right: Illustration of our Direct Regression, Sparse Correspondence and Dense Correspondence models.}
    \label{fig:all-three-methods}
    \vspace{-0.5cm}
\end{figure}

In this section, we present three methods that are conceptually promising for eye-in-hand camera calibration from a single RGB image, and which also raised interesting and novel scientific questions that we discuss in sections \ref{sec:result} and \ref{sec:analysis}. The first method is an end-to-end direct regression method that estimates the camera's pose from an RGB image. In contrast, the other two methods use deep learning to provide key missing components for well-established geometric pose estimation approaches. The latter have been shown to work well in other problem settings~\citep{lee2020camera, DPOD, sheffer2020pnp,xiong2021region}, but eye-in-hand camera calibration has the particularity that not only the camera is too close to the end-effector for typical vision sensors to capture a depth image of it, but also the visible geometry varies widely within our inference space. To test whether it is still feasible to use such approaches, our second method uses deep learning to regress the 2D locations of predefined keypoints on the end-effector, and passes them to the PnP ~\cite{PnP} algorithm to solve for the camera's pose. The main challenge with this method is that the large variations in the visible geometry lead to keypoint occlusions and to some keypoints being often outside of the image frame. This is illustrated in fig.~\ref{fig:summary}. Our final method tests whether we can alleviate this issue by using deep learning to regress a segmented depth image of the end-effector and an initial guess of the camera's pose, and the Iterative Closest Point (ICP) \citep{point-to-plane-icp, point-to-point-icp, original-icp-paper-3} algorithm to refine this initial estimate. All three methods are illustrated in fig.~\ref{fig:all-three-methods}.
%
%
\subsection{Problem Setting}
%
%
Our problem setting consists of the typical eye-in-hand camera calibration problem, where the aim is to estimate the camera's extrinsic matrix, which is the camera to end-effector pose, $T_{EC}~=~[R_{EC}|t_{ TC}]~\in~SE(3)$, where $R_{EC}\in SO(3)$ and $t_{EC}\in \mathds{R}^3$ is the orientation and position of the camera in the end-effector frame respectively. We also define the end-effector to robot base pose as $T_{BE}$, the calibration object to camera pose as $T_{CO}$, and an image captured by the wrist-mounted camera as $I$. Throughout this paper, we use $\Tilde{\cdot}$ to denote an estimated quantity.

As opposed to classical approaches, we constrain ourselves to eye-in-hand camera calibration from a single RGB image without any external apparatus, enabling our methods to be readily deployed in the wild. We further constrain ourselves to using only synthetic data in order to alleviate the costly data requirements of deep learning approaches, and to obtain ground truth labels that facilitate effective learning. Finally, we assume that we have an accurate estimate of the camera intrinsic matrix, which the manufacturer typically provides, and access to the CAD model of our robot's end-effector.
%
%
\subsection{Direct Regression Model}\label{method:direct-regression}
%
%
Direct regression is the end-to-end deep learning method we consider. As illustrated in the left diagram of fig.~\ref{fig:all-three-methods}, a neural network (NN) is tasked with regressing the camera's pose from a single RGB image and is trained in simulation using ground truth labels in a supervised setting. The only architectural constraint we introduce comes from the parameterisation of the orientation, where we use the  6D rotation encoding introduced in \citep{rotation-encoding}. 

Overall, our network outputs a $9$ dimensional tensor, with 3 dimensions representing position and 6 the orientation. It consists of a convolutional encoder followed by linear layers that regress the camera pose. We trained it with the ground truth labels using the Mean Squared Error (MSE) loss, and we fully breakdown the architecture and training details in Appendix A.  
%
%
\subsection{Sparse Correspondence Model}\label{method:pnp}

\begin{figure}
    \centering
    \includegraphics[width=\linewidth]{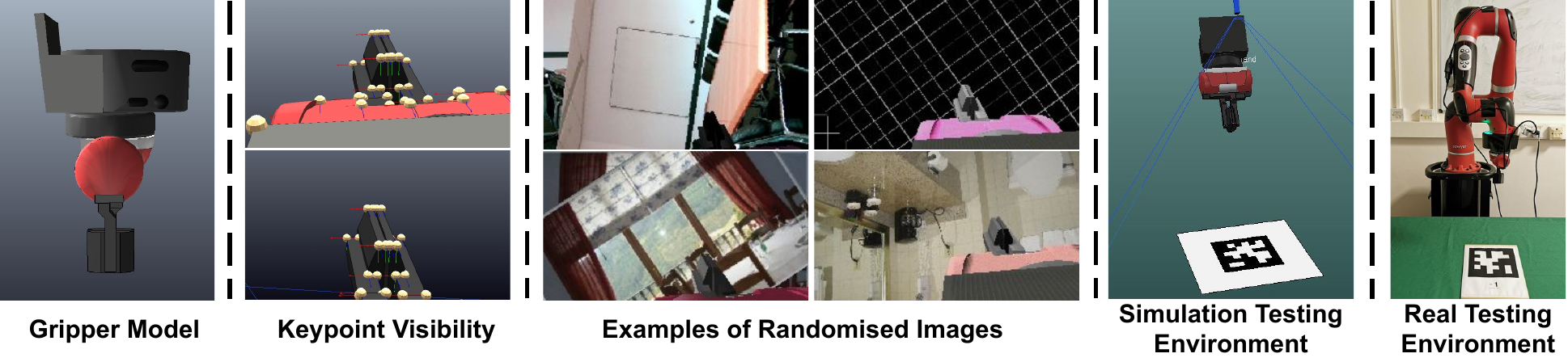}
    \caption{Left to right: Model of the gripper. Illustration of keypoint visibility from different viewpoints. Examples of randomised simulation images. Simulation and real world testing environments.}
    \label{fig:summary}
    \vspace{-0.4cm}
\end{figure}

Our sparse correspondence model is illustrated in the middle of fig.~\ref{fig:all-three-methods}. Starting from an RGB image, it uses a U-Net~\citep{unet} type architecture with a spatial-soft-argmax~\citep{finn2016deep} output activation in order to regress a set of 2D keypoints in image space. We use the sum of an L1 and MSE losses to train our network on the ground truth labels, such that each output dimension predicts the 2D location of a predefined 3D keypoint. A detailed breakdown of our architecture and training parameters can be found in Appendix A.

At test time, we use the network's output, the corresponding 3D keypoint locations defined in the object's model, and the camera's intrinsic matrix to estimate the camera's pose with PnP and RANSAC~\citep{PnP}, which are a popular pose estimation and outlier rejection algorithms.

Finally, in order to alleviate the issue of some keypoints being out of frame depending on the viewpoint, we (1) consider 38 keypoints that are present in at least 70\% of the training images, and (2) for any keypoint that remains out of frame, we train the model to predict its projection in the image plane. This way, at test time, we can infer which keypoints are out of frame by looking at the predictions at the rim, and ignoring these predictions for PnP pose estimation.
%
%
\subsection{Dense Correspondence Model}\label{method:icp}
%
%
Our Dense Correspondence model is illustrated on the right of fig. \ref{fig:all-three-methods}. Starting from an RGB image, it uses three independently trained NNs to estimate the camera's pose in two stages.  In the first stage, the three NNs are used to predict (1) a depth map, (2) a segmentation mask of the gripper, and (3) an ICP initialisation. In the second stage, (1) the segmentation mask is used to segment the depth map to only return depth values on gripper pixels, (2) the camera intrinsic matrix is used to project the segmented depth map to a point cloud, and (3) ICP is used to establish dense correspondences between the estimated point cloud and the model of the end-effector and to refine the camera's pose estimate from the first stage. The network used for ICP initialisation is the same as the one used for direct regression. The same U-Net architecture as for the sparse correspondence model is used to predict the depth and segmentation maps. See Appendix A for further architecture design choices and training details.
%
%
\subsection{Fusing Multiple Estimates}\label{sec:method:outlier-rejection}
%
%
Compared to classical eye-in-hand camera calibration methods, which require calibration to be done offline and with special apparatus, the proposed deep learning alternatives can estimate the camera's pose online in unstructured environments. This creates the possibility of fusing multiple estimates to increase the overall calibration accuracy. In order to explore this, we implemented a simple aggregation algorithm that rejects 20\% of the least likely samples from a set of candidate estimates under a Gaussian data model. It then averages the remaining samples together to yield the final estimate. Our algorithm for this procedure is available in Appendix C.
%
%
\subsection{Dataset Generation and Sim-to-Real Transfer}

We train our networks entirely with simulated data using the Coppelia~\cite{coppeliaSim} simulator. In order to generate the dataset, we randomise the pose of the camera relative to the end-effector at each simulated timestep and record (1) the current RGB, depth and end-effector segmentation images, (2) the ground truth extrinsic matrix, and (3) the 2D keypoint locations in image space. 

In order to overcome the ``reality gap'', we apply visual domain randomisation~\cite{valassakis2020crossing, tobin2017domain,alghonaim2021benchmarking}. Specifically,  we (1) randomise the colours/textures of the simulated gripper and the light sources on the simulator, (2) replace the background of our images with random images of textures and indoor scenes~\cite{quattoni2009recognizing, finn2017one}, and (3) apply a post-processing colour jitter operation~\citep{NEURIPS2019_9015}, further randomly perturbing the brightness, contrast, saturation and hue of the whole image. In total, we generated $10~000$ labelled images in approximately 30 minutes, examples of which can be seen in fig.~\ref{fig:summary}.
%
%
\section{Experiments}\label{sec:result}
%
%
In our experiments, we evaluate how the different proposed deep learning approaches compare to each other and to various standard off-the-shelf calibration methods both in simulation and in the real world. From them emerges the surprising result that simple direct regression using end-to-end deep learning outperforms both the classical approaches tested and the correspondence-based deep learning methods. As such, we perform a series of analysis experiments, which we describe in section~\ref{sec:analysis}, that are aimed at gaining insights into this surprising observation. 

In our comparisons, we include all our deep learning-based methods and the following established methods readily available on OpenvCV~\citep{opencv_library}: Dainiilidis et al.~\citep{daniilidis1999hand}, Tsai and Lenz~\citep{tsai1989new}, Doraika and Harod~\citep{horaud1995hand}, and  Park and Martin~\citep{park1994robot}. For all experiments, we use the Realsense D435 (or its simulated counterpart) at a $480\times848$ resolution, and for the deep learning methods we downsample these images to a resolution of $144\times256$.

Comparing classical methods to our deep learning ones fairly can be somewhat challenging since classical methods require a training set of several end-effector to robot and calibration object to camera poses to perform a single calibration, while our methods only require a single image. In sections \ref{sim-experiment} and \ref{sec:real-experiment} we describe our procedures and what we did to medicate this issue, and in Appendix D we provide our algorithm for each of these procedures.
%
%
\subsection{Simulation Experiment}\label{sim-experiment}
%
%
Our simulated environment consists of a wrist-mounted camera attached to a Sawyer robot's gripper that is free to move around an AprilTag (see fig.~\ref{fig:summary}). To benchmark all methods, we (1) collect a dataset $\{I_i, T_{BE}^i, \Tilde{T}_{CO}^i\}_{i=1}^{15}$ of images and corresponding AprilTag and end-effector poses, with the tag poses $\Tilde{T}_{CO}$ estimated using the AprilTags3 library~\citep{april-tag-sim-experiment,Wang2016,malyuta:2017mt}, (2) use all of the $15$ datapoints in this dataset to estimate the extrinsic matrix using each classical method, and evaluate these estimates, (3) use each of the $15$ datapoints in this dataset independently to estimate the extrinsic matrix using each of the proposed learned methods, and evaluate each of the single image estimates, and (4) for each of the learned methods, we fuse together all 15 independent predictions using our fusion procedure described in~\ref{sec:method:outlier-rejection} and evaluate the fused estimate. 

We repeat this procedure for 100 different ground truth extrinsic matrices, take the average and standard deviation, and display the results in table~\ref{tab:main_results}. We evaluate translation and rotation independently: Given an estimate of the extrinsics $\Tilde{T}_{EC}=[\Tilde{R}_{EC}|\Tilde{t}_{EC}]$, with ground truth $T_{EC}^*=[R_{EC}^*|t_{EC}^*]$, we define the position error as $e_t = ||\Tilde{t}_{EC} - t_{EC}^*||_2$ , where $||\cdot||_2$ is the L2 norm, and the rotational error as $e_R = \theta$, the angle from the axis-angle representation of the rotation matrix $R_{\Delta}=(w, \theta)$ that satisfies the relationship $R_{EC}^*=R_{\Delta}\Tilde{R}_{EC}$, where $w$ and $\theta$ are the axis and angle of rotation. 
%
%
\subsection{Real World Experiment}\label{sec:real-experiment}
%
%
The real-world evaluation environment is analogous to the simulated environment and consists of a Sawyer robot with a wrist-mounted camera moving around an AprilTag (see fig.~\ref{fig:summary}). In order to evaluate our methods in the real world, we (1) collect a training data bank $\mathcal{D}_{train}=\{I_i, T_{BE}^i, \Tilde{T}_{CO}^i\}_{i=1}^{40}$ and an evaluation dataset $\mathcal{D}_{eval}=\{ T_{BE}^i, \Tilde{T}_{CO}^i\}_{i=1}^{60}$ automatically by scripting a trajectory around the AprilTag, where AprilTag poses are estimated using the AprilTags3 library~\citep{april-tag-sim-experiment,Wang2016,malyuta:2017mt}, (2) we sample $40$ training datasets of $15$ datapoints each from our training data bank, (3) we use all $15$ datapoints in each training set to get an estimate of the camera extrinsic matrix using each classical method, and evaluate that estimate, (4) we use each one of the $15$ datapoints in each dataset to get an estimate of the extrinsic matrix using our learned methods, and evaluate each single estimate, and (5) for each of the learned methods, we fuse together all 15 independent predictions using the procedure described in~\ref{sec:method:outlier-rejection} and evaluate the fused estimate.

We repeat this procedure for two different ground truth extrinsics, calculate the average and standard deviation, and display the results in table~\ref{tab:main_results}. Since ground truth extrinsic parameters are not available in the real world, we take inspiration from~\cite{daniilidis1999hand,tsai1989new} and use an indirect error metric: For each corresponding end-effector to robot pose and calibration object to camera pose from the evaluation dataset, $\{T_{BE}^i, \Tilde{T}_{CO}^i\}\in \mathcal{D}_{eval}$, we estimate the pose of the calibration object in the robot's frame, $\Tilde{T}_{BO}^i=T_{BE}^i \Tilde{T}_{EC} \Tilde{T}_{CO}^i=[\Tilde{R}_{WO}^i|\Tilde{T}_{BO}^i]$, where $\Tilde{T}_{EC}$ is the estimate of the camera to end-effector pose that we are evaluating. We then define the error metric $\epsilon_{std} = \left(1/60 \sum_{i=1}^{60} \left||\Tilde{t}_{BO}^i - \mu \right||^2_2\right)^{1/2}$ as the standard deviation of the estimated calibration object position, where $\mu=1/60 \sum_{i=1}^{60} \Tilde{t}_{BO}^i$ is the estimated mean object position.

We emphasise that although the AprilTag is visible in all images in both the simulated and real environments, it is never used to help the calibration of our deep learning methods. We use these images simply to ensure that we have the exact same calibration and evaluation datasets for both classical and our learned methods. 

\begin{table}
\small
    \centering
        \begin{tabular}{|l|c|c|c|}
        \toprule
         & \multicolumn{2}{|c|}{Simulation} & Real World \\
        \toprule
        Method &      $\epsilon_t$ [mm] & $\epsilon_R$ [degrees] & $\epsilon_{std}$ [mm] \\
        \midrule
        TSAI~\citep{tsai1989new}    &  $(171.4 \pm 271.2)$ &      $(15.9 \pm 20.6)$ & $(17.5 \pm 18.9)$\\
        PARK~\citep{park1994robot}                         &    $(68.7 \pm 70.1)$ &        $(6.3 \pm 4.7)$ & $(15.3 \pm 14.3)$\\
        HORAUD~\citep{horaud1995hand}                       &    $(68.9 \pm 72.2)$ &        $(6.3 \pm 4.7)$ & $(16.3 \pm 16.6)$\\
        DANIILIDIS~\citep{daniilidis1999hand}                   &  $(110.8 \pm 122.8)$ &      $(11.5 \pm 29.4)$ & $(14.7 \pm 13.6)$\\
        DR                     &     $\boldsymbol{(13.4 \pm 4.1)}$ &        $\boldsymbol{(4.4 \pm 1.4)}$ & $(10.6 \pm 4.1)$\\
        DR (fusion)             &     $\boldsymbol{(13.4 \pm 4.1)}$ &        $\boldsymbol{(4.4 \pm 1.4)}$ &  $\boldsymbol{(10.4 \pm 4.0)}$\\
        SC        &  $(363.6 \pm 353.1)$ &     $(117.4 \pm 48.2)$ &  $(516.4 \pm 338.1)$ \\
        SC (fusion) &  $(338.5 \pm 301.0)$ &     $(120.2 \pm 42.7)$ & $(291.2 \pm 129.6)$ \\
        DC             &    $(93.5 \pm 45.9)$ &      $(33.9 \pm 25.7)$ & $(21.0 \pm 10.4)$ \\
        DC (fusion)    &    $(73.4 \pm 21.2)$ &      $(23.1 \pm 13.9)$ &  $(15.0 \pm 1.7)$ \\
        \bottomrule
        \end{tabular}
    \caption{Evaluation of the classical methods, our Direct Regression (DR), Sparse Correspondence (SC) and Dense Correspondence (DC) methods, and their fusion variants with aggregated estimates.}
    \vspace{-0.3cm}
    \label{tab:main_results}
\end{table}
%
%
\subsection{Results}\label{results}
%
%
All benchmarking results are shown in table \ref{tab:main_results}. Our main observation is that the direct regression method outperforms all others, which is a surprising result. First, it was not expected to outperform the classical methods since they use analytical solutions to estimate the extrinsics, which in the absence of noise in the system should give a perfect calibration. We believe this stems from using automatic data gathering for the classical methods, which does not allow for carefully curating the calibration dataset and the distribution of poses within it, a manual trial-and-error process that is generally required to bolster calibration accuracy. This also raises the question about the sensitivity of these methods to said noise, which we investigate in section~\ref{sec:analysis}. 

Second, it was also not expected to outperform the correspondence-based learned methods, since those introduce strong geometric constraints that intuitively should help guide the model to good solutions. We thoroughly investigate the reasons behind this observation in sections~\ref{analysis:sc} and~\ref{analysis:dc}.

\vspace{-0.3cm}
%
%
\section{Analysis}\label{sec:analysis}
%
%
In this section, we aim to build a deeper understanding of the surprising results that we observed in our experiments, which showed that a direct regression of calibration parameters outperforms methods that incorporate well-understood geometric modelling. We split this section into three parts, each analysing the performance of one particular approach.

\vspace{-0.2cm}
%
%
\subsection{Classical Methods}
%
%
\begin{wrapfigure}{r}{0.4\linewidth}
\vspace{-1.0cm}
\centering
\includegraphics[width = \linewidth]{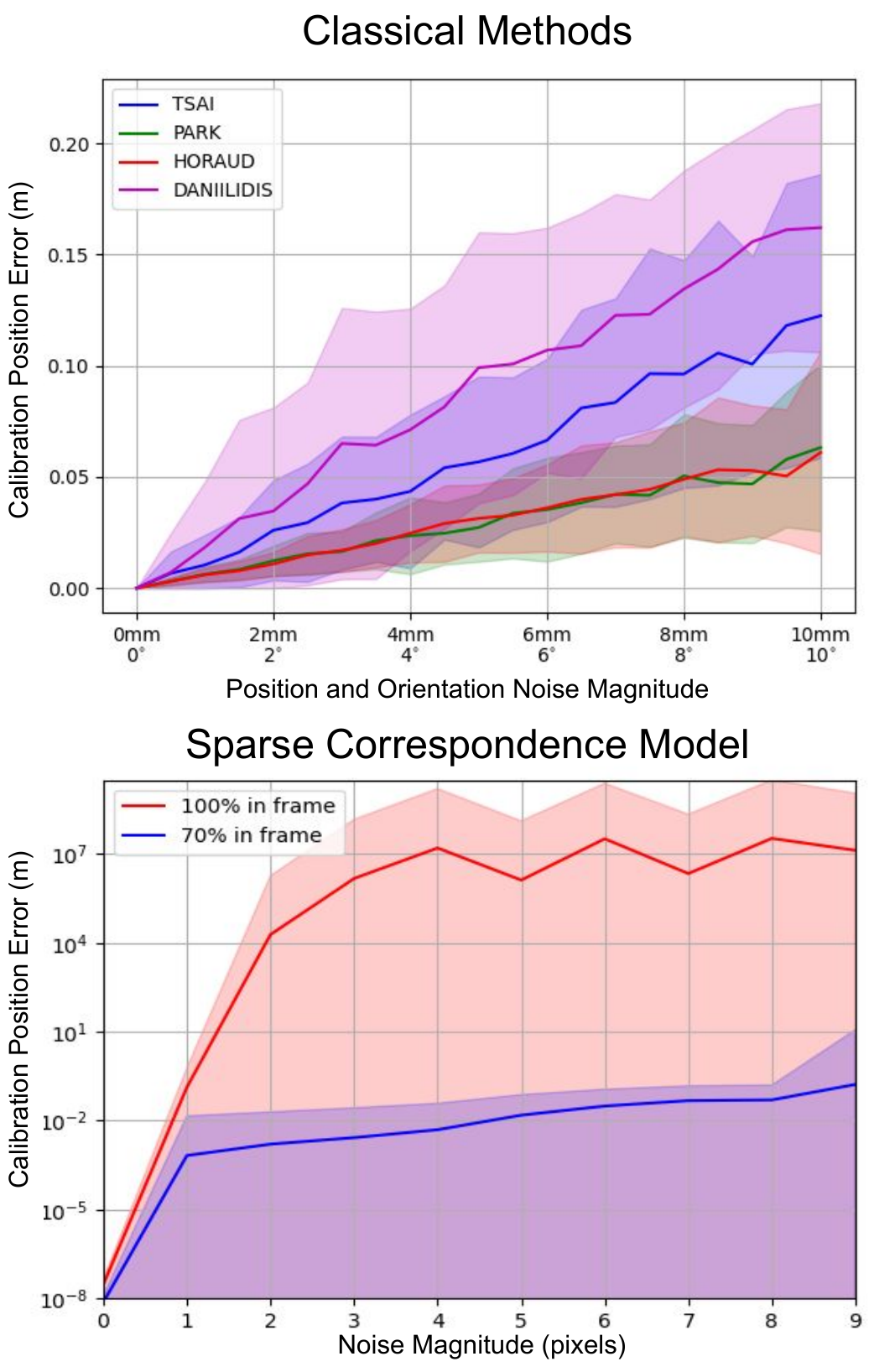}
\caption{Top: Analysis of sensitivity to noise in the estimated tag positions of classical methods. Bottom: Analysis of sensitivity to noise in the estimated keypoint positions of PnP+RANSAC.}
\vspace{-1.0cm}
\label{fig:noise-analysis}
\end{wrapfigure}

To better understand the results from our experiments in section~\ref{sec:result}, we first aim to assess how sensitive classical methods are to sources of noise in the system. Assuming the main source of error stems from the calibration object pose estimation, we fix every other source of noise in simulation to their ground truth values. We then vary the noise injected to the calibration object's poses by controlled amounts and observe how this affects the quality of the calibration result from classical methods.

Precisely, we vary the noise levels from $0mm$ to $10mm$ and $0^{\circ}$ to $10^{\circ}$, in increments of $0.5mm$ and $0.5^{\circ}$, and for each noise tier, we perform calibration with each of the classical methods considered. We then compare the result to the ground truth extrinsics and obtain position errors. We repeat this procedure for 100 camera extrinsics, with the average errors obtained and their standard deviations illustrated in the top graph of fig.~\ref{fig:noise-analysis}.

We can see that, as expected, with perfect information, the classical methods return an exact solution. However, we also see that the calibration quality rapidly deteriorates with increased noise in the calibration object's poses. This illustrates our motivation for using a deep learning-based approach, since our methods are independent of any test-time data gathering or calibration object pose estimation quality.

\vspace{-0.2cm}
%
%
\subsection{Sparse Correspondences}\label{analysis:sc}
%
%
In order to better understand why our sparse correspondence model did not perform strongly we set up another controlled experiment. Starting from ground truth 2D-3D correspondences, we add fixed amounts of noise to the 2D keypoint locations and observe the effect of this on the quality of PnP + RANSAC pose estimation. Specifically, we iterate through $200$ random extrinsic matrices, and for each \{2D keypoints, 3D keypoints, extrinsic\} tuple, we perform PnP + RANSAC with the noise-injected 2D keypoints and compare the result to the ground truth. We repeat this for different magnitudes of noise, ranging from $0$ to $9$ pixels in increments of $1$ pixel, and averaged over all the datapoints considered. We plot the resulting errors in the "70\% in frame" curve in the bottom graph of fig.~\ref{fig:noise-analysis}.

During our investigation we also considered using a smaller number of keypoints, but ones that always remain in frame. With those we observed that even though this is an easier task for the network prediction, our final extrinsics estimation did not improve. In order to understand this behaviour, we repeat the controlled noise experiment but only considering the keypoints that always appear in frame in the images, which resulted in $12$ remaining keypoints clustered on the gripper's fingers. This is plotted in the "100\% in frame" curve in the bottom graph of fig.~\ref{fig:noise-analysis}.

From fig.~\ref{fig:noise-analysis}, we clearly see that there is a very high sensitivity of sparse correspondence pose estimation to errors in 2D keypoint pixel locations. For the keypoints in our training set, to get an extrinsic calibration position error of less than 1cm, our networks' average keypoint prediction error would need to be less than 4 to 5 pixels. This is with the additional challenge of the network having to keep consistent predictions even though sometimes the keypoints appear out of frame. On the other hand, if we consider the easier to learn problem of only predicting keypoints that always appear in-frame, the sensitivity skyrockets after single-pixel average error in 2D keypoint locations. Overall, we believe that this sensitivity to noise is the primary reason for why the sparse correspondence model did not give a strong performance.
%
%
\subsection{Dense  Correspondences}\label{analysis:dc}
%
%
\begin{figure}
    \centering
    \includegraphics[width = \linewidth]{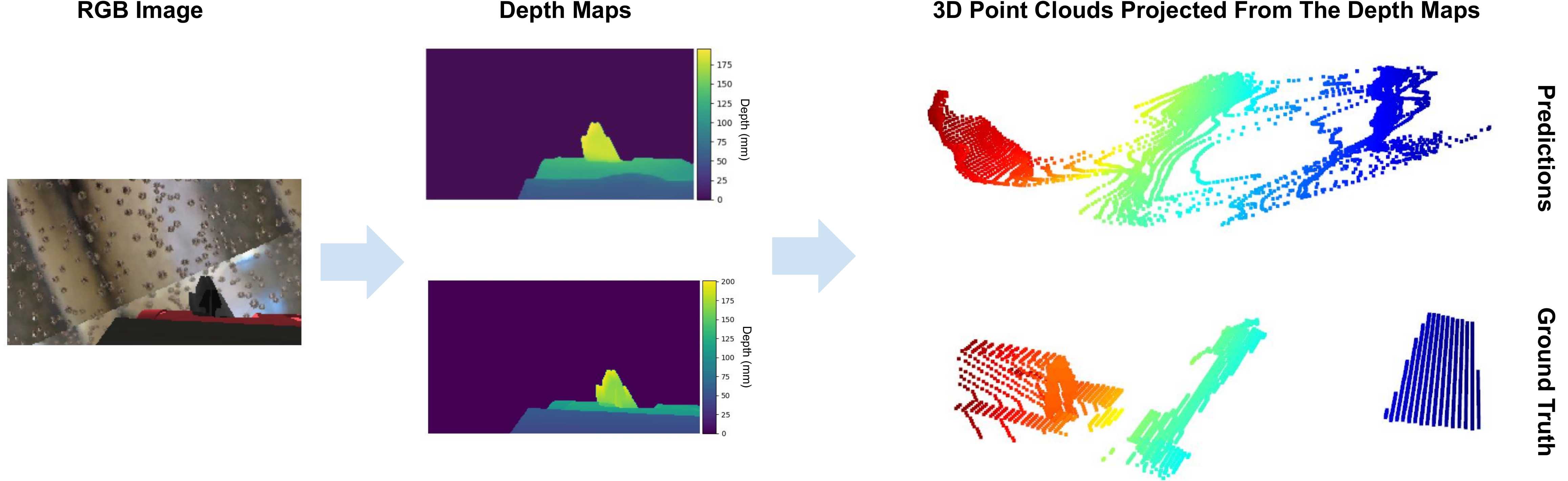}
    \caption{Left: RGB image of the gripper rendered in simulation. Top centre: Depth image predicted by our model from the simulated RGB image. Top right: point cloud projected from the predicted depth image. Bottom centre: Depth image rendered in simulation from the same viewpoint.  Bottom right, point cloud projected from the ground truth depth image.}
    \label{fig:dense_analysis}
    \vspace{-0.2cm}
\end{figure}

Although conceptually using a dense geometric correspondence approach should alleviate some of the drawbacks of using sparse correspondences, we found that in practice our model that uses ICP to refine estimates made by the direct regression method decreases their accuracy.

The reason for this becomes apparent when considering the illustrations shown in fig.~\ref{fig:dense_analysis}. Although the depth values obtained by the neural networks look appealing when projected onto the image plane, they fall short when considering the full 3D structure. We conjecture that the main reason for this is that in order to recover a correct 3D structure from such close distances, depth values need to have strong discontinuous jumps that typical neural networks have difficulty modelling. This is supported by the observations in fig.~\ref{fig:dense_analysis}: We can see that the overall depth increase seems to match the ground truth values, but there is an averaging effect that tends to make the depth values increase smoothly, which does not allow us to recover the correct shape. As such, we can conclude that using a simple depth regression technique in this setting is inappropriate for this task's requirements, with further research needed to overcome the discontinuity problem.

\vspace{-0.2cm}
%
%
\section{Conclusion}\label{sec:conclusion}
%
%
In this work, we presented and evaluated three deep learning-based methods for online eye-in-hand camera calibration from a single image. We trained all our models entirely on synthetic images, and evaluated them against each other and long-established calibration methods in simulation and the real world. Surprisingly, our experiments indicated that a direct regression method from images to camera extrinsics outperformed other alternatives. In order to better understand this result, we then conducted a series of introspection experiments, which indicated that a strong shortcoming of geometry-based methods seems to be their sensitivity to noise in their input data. Finally, while depth map regression for dense correspondences conceptually could have provided the answer to this, our experiments indicated that due to pronounced discontinuities in the depth profiles of such close-up viewpoints, further research is needed, specifically in learning depth prediction for eye-in-hand robot manipulation.


\clearpage

\acknowledgments{This work was supported by the Royal Academy of Engineering under the Research Fellowship scheme.}


\bibliography{references.bib}  

\begin{thebibliography}{48}
\providecommand{\natexlab}[1]{#1}
\providecommand{\url}[1]{\texttt{#1}}
\expandafter\ifx\csname urlstyle\endcsname\relax
  \providecommand{\doi}[1]{doi: #1}\else
  \providecommand{\doi}{doi: \begingroup \urlstyle{rm}\Url}\fi

\bibitem[Johns(2021)]{johns2021coarse}
E.~Johns.
\newblock {Coarse-to-Fine Imitation Learning: Robot Manipulation from a Single
  Demonstration}.
\newblock In \emph{IEEE International Conference on Robotics and Automation
  (ICRA)}, 2021.

\bibitem[Litvak et~al.(2019)Litvak, Biess, and
  Bar-Hillel]{cam-cali-for-manipulation}
Y.~Litvak, A.~Biess, and A.~Bar-Hillel.
\newblock {Learning Pose Estimation for High-Precision Robotic Assembly Using
  Simulated Depth Images}.
\newblock In \emph{IEEE International Conference on Robotics and Automation
  (ICRA)}, 2019.

\bibitem[Johns et~al.(2016)Johns, Leutenegger, and Davison]{johns2016deep}
E.~Johns, S.~Leutenegger, and A.~J. Davison.
\newblock {Deep Learning a Grasp Function for Grasping under Gripper Pose
  Uncertainty}.
\newblock In \emph{IEEE International Conference on Intelligent Robots and
  Systems (IROS)}, 2016.

\bibitem[Valassakis et~al.(2021)Valassakis, Di~Palo, and
  Johns]{valassakis2021coarse}
E.~Valassakis, N.~Di~Palo, and E.~Johns.
\newblock {Coarse-to-Fine for Sim-to-Real: Sub-Millimetre Precision Across Wide
  Task Spaces}.
\newblock In \emph{IEEE International Conference on Intelligent Robots and
  Systems (IROS)}, 2021.

\bibitem[Puang et~al.(2020)Puang, Peng~Tee, and Jing]{puang2020kovis}
E.~Y. Puang, K.~Peng~Tee, and W.~Jing.
\newblock {KOVIS: Keypoint-based Visual Servoing with Zero-Shot Sim-to-Real
  Transfer for Robotics Manipulation}.
\newblock In \emph{IEEE International Conference on Intelligent Robots and
  Systems (IROS)}, 2020.

\bibitem[Daniilidis(1999)]{daniilidis1999hand}
K.~Daniilidis.
\newblock {Hand-Eye Calibration Using Dual Quaternions}.
\newblock \emph{The International Journal of Robotics Research (IJRR)}, 1999.

\bibitem[Tsai et~al.(1989)Tsai, Lenz, et~al.]{tsai1989new}
R.~Y. Tsai, R.~K. Lenz, et~al.
\newblock {A new technique for fully autonomous and efficient 3D robotics
  hand/eye calibration}.
\newblock \emph{IEEE Transactions on Robotics and Automation}, 1989.

\bibitem[Park and Martin(1994)]{park1994robot}
F.~C. Park and B.~J. Martin.
\newblock {Robot sensor calibration: solving AX= XB on the Euclidean group}.
\newblock \emph{IEEE Transactions on Robotics and Automation}, 1994.

\bibitem[Horaud and Dornaika(1995)]{horaud1995hand}
R.~Horaud and F.~Dornaika.
\newblock {Hand-eye Calibration}.
\newblock \emph{The International Journal of Robotics Research (IJRR)}, 1995.

\bibitem[{Labb\'e} et~al.(2021){Labb\'e}, {Carpentier}, {Aubry}, and
  {Sivic}]{labbe2021single}
Y.~{Labb\'e}, J.~{Carpentier}, M.~{Aubry}, and J.~{Sivic}.
\newblock {Single-view robot pose and joint angle estimation via render \&
  compare}.
\newblock In \emph{Conference on Computer Vision and Pattern Recognition
  (CVPR)}, 2021.

\bibitem[Lee et~al.(2020)Lee, Tremblay, To, Cheng, Mosier, Kroemer, Fox, and
  Birchfield]{lee2020camera}
T.~E. Lee, J.~Tremblay, T.~To, J.~Cheng, T.~Mosier, O.~Kroemer, D.~Fox, and
  S.~Birchfield.
\newblock {Camera-to-Robot Pose Estimation from a Single Image}.
\newblock In \emph{IEEE International Conference on Robotics and Automation
  (ICRA)}, 2020.

\bibitem[Shiu and Ahmad(1989)]{shiu1987calibration}
Y.~C. Shiu and S.~Ahmad.
\newblock {Calibration of wrist-mounted robotic sensors by solving homogeneous
  transform equations of the form AX=XB}.
\newblock \emph{IEEE Transactions on Robotics and Automation}, 1989.

\bibitem[Andreff et~al.(2001)Andreff, Horaud, and Espiau]{andreff2001robot}
N.~Andreff, R.~Horaud, and B.~Espiau.
\newblock {Robot Hand-Eye Calibration using Structure from Motion}.
\newblock \emph{The International Journal of Robotics Research (IJRR)}, 2001.

\bibitem[Pauwels and Kragic(2016)]{pauwels2016integrated}
K.~Pauwels and D.~Kragic.
\newblock Integrated on-line robot-camera calibration and object pose
  estimation.
\newblock In \emph{IEEE International Conference on Robotics and Automation
  (ICRA)}, 2016.

\bibitem[Furrer et~al.(2018)Furrer, Fehr, Novkovic, Sommer, Gilitschenski, and
  Siegwart]{furrer2018evaluation}
F.~Furrer, M.~Fehr, T.~Novkovic, H.~Sommer, I.~Gilitschenski, and R.~Siegwart.
\newblock {Evaluation of Combined Time-Offset Estimation and Hand-Eye
  Calibration on Robotic Datasets}.
\newblock In \emph{Field and Service Robotics (FSR)}, 2018.

\bibitem[Faion et~al.(2012)Faion, Ruoff, Zea, and Hanebeck]{faion2012recursive}
F.~Faion, P.~Ruoff, A.~Zea, and U.~D. Hanebeck.
\newblock {Recursive Bayesian calibration of depth sensors with non-overlapping
  views}.
\newblock In \emph{International Conference on Information Fusion (FUSION)},
  2012.

\bibitem[Antonello et~al.(2017)Antonello, Gobbi, Michieletto, Ghidoni, and
  Menegatti]{antonello2017fully}
M.~Antonello, A.~Gobbi, S.~Michieletto, S.~Ghidoni, and E.~Menegatti.
\newblock {A fully automatic hand-eye calibration system}.
\newblock In \emph{European Conference on Mobile Robots (ECMR)}, 2017.

\bibitem[Liu et~al.(2020)Liu, Madhusudanan, Chen, Li, Ge, Ru, and
  Sun]{liu2020fast}
X.~Liu, H.~Madhusudanan, W.~Chen, D.~Li, J.~Ge, C.~Ru, and Y.~Sun.
\newblock {Fast Eye-in-Hand 3-D Scanner-Robot Calibration for Low Stitching
  Errors}.
\newblock \emph{IEEE Transactions on Industrial Electronics}, 2020.

\bibitem[Li et~al.(2018)Li, Wang, Ji, Xiang, and Fox]{li2018deepim}
Y.~Li, G.~Wang, X.~Ji, Y.~Xiang, and D.~Fox.
\newblock {DeepIM: Deep Iterative Matching for 6D Pose Estimation}.
\newblock In \emph{European Conference on Computer Vision (ECCV)}, 2018.

\bibitem[Labb{\'e} et~al.(2020)Labb{\'e}, Carpentier, Aubry, and
  Sivic]{labbe2020cosypose}
Y.~Labb{\'e}, J.~Carpentier, M.~Aubry, and J.~Sivic.
\newblock {CosyPose: Consistent multi-view multi-object 6D pose estimation}.
\newblock In \emph{European Conference on Computer Vision (ECCV)}, 2020.

\bibitem[Valassakis et~al.(2020)Valassakis, Ding, and
  Johns]{valassakis2020crossing}
E.~Valassakis, Z.~Ding, and E.~Johns.
\newblock {Crossing the Gap: A Deep Dive into Zero-Shot Sim-to-Real Transfer
  for Dynamics}.
\newblock In \emph{IEEE International Conference on Intelligent Robots and
  Systems (IROS)}, 2020.

\bibitem[Tobin et~al.(2017)Tobin, Fong, Ray, Schneider, Zaremba, and
  Abbeel]{tobin2017domain}
J.~Tobin, R.~Fong, A.~Ray, J.~Schneider, W.~Zaremba, and P.~Abbeel.
\newblock {Domain Randomization for Transferring Deep Neural Networks from
  Simulation to the Real World}.
\newblock In \emph{IEEE International Conference on Intelligent Robots and
  Systems (IROS)}, 2017.

\bibitem[Alghonaim and Johns(2021)]{alghonaim2021benchmarking}
R.~Alghonaim and E.~Johns.
\newblock {Benchmarking Domain Randomisation for Visual Sim-to-Real Transfer}.
\newblock In \emph{IEEE International Conference on Robotics and Automation
  (ICRA)}, 2021.

\bibitem[Fischler and Bolles(1981)]{PnP}
M.~A. Fischler and R.~C. Bolles.
\newblock {Random Sample Consensus: A Paradigm for Model Fitting with
  Applications to Image Analysis and Automated Cartography}.
\newblock \emph{Association for Computing Machinery (ACM)}, 1981.

\bibitem[Haugaard et~al.(2020)Haugaard, Langaa, Sloth, and
  Buch]{haugaard2020fast}
R.~L. Haugaard, J.~Langaa, C.~Sloth, and A.~G. Buch.
\newblock Fast robust peg-in-hole insertion with continuous visual servoing.
\newblock \emph{arXiv e-prints}, 2020.

\bibitem[Zakharov et~al.(2019)Zakharov, Shugurov, and Ilic]{DPOD}
S.~Zakharov, I.~Shugurov, and S.~Ilic.
\newblock {DPOD: 6D Pose Object Detector and Refiner}.
\newblock In \emph{IEEE International Conference on Computer Vision (ICCV)},
  2019.

\bibitem[Sheffer and Wiesel(2020)]{sheffer2020pnp}
R.~Sheffer and A.~Wiesel.
\newblock {PnP-Net: A hybrid Perspective-n-Point Network}.
\newblock \emph{arXiv e-prints}, 2020.

\bibitem[Xiong et~al.(2021)Xiong, Liu, and Chen]{xiong2021region}
F.~Xiong, C.~Liu, and Q.~Chen.
\newblock {Region Pixel Voting Network (RPVNet) for 6D Pose Estimation from
  Monocular Image}.
\newblock \emph{Applied Sciences}, 2021.

\bibitem[Chen and Medioni(1991)]{point-to-plane-icp}
Y.~Chen and G.~Medioni.
\newblock Object modeling by registration of multiple range images.
\newblock In \emph{IEEE International Conference on Robotics and Automation
  (ICRA)}, 1991.

\bibitem[Besl and McKay(1992)]{point-to-point-icp}
P.~Besl and N.~D. McKay.
\newblock {a method for registration of 3-D shapes}.
\newblock \emph{IEEE Transactions on Pattern Analysis and Machine Intelligence
  (TPAMI)}, 1992.

\bibitem[Zhang(1994)]{original-icp-paper-3}
Z.~Zhang.
\newblock Iterative point matching for registration of free-form curves and
  surfaces.
\newblock \emph{International Journal of Computer Vision (IJCV)}, 1994.

\bibitem[Zhou et~al.(2019)Zhou, Barnes, Lu, Yang, and Li]{rotation-encoding}
Y.~Zhou, C.~Barnes, J.~Lu, J.~Yang, and H.~Li.
\newblock {On the Continuity of Rotation Representations in Neural Networks}.
\newblock In \emph{IEEE Conference on Computer Vision and Pattern Recognition
  (CVPR)}, 2019.

\bibitem[Ronneberger et~al.(2015)Ronneberger, Fischer, and Brox]{unet}
O.~Ronneberger, P.~Fischer, and T.~Brox.
\newblock {U-Net: Convolutional Networks for Biomedical Image Segmentation}.
\newblock In \emph{International Conference on Medical Image Computing and
  Computer Assisted Intervention (MICCAI)}, 2015.

\bibitem[Finn et~al.(2016)Finn, Tan, Duan, Darrell, Levine, and
  Abbeel]{finn2016deep}
C.~Finn, X.~Y. Tan, Y.~Duan, T.~Darrell, S.~Levine, and P.~Abbeel.
\newblock {Deep Spatial Autoencoders for Visuomotor Learning}.
\newblock In \emph{IEEE International Conference on Robotics and Automation
  (ICRA)}, 2016.

\bibitem[Rohmer et~al.(2013)Rohmer, Singh, and Freese]{coppeliaSim}
E.~Rohmer, S.~P.~N. Singh, and M.~Freese.
\newblock {CoppeliaSim (formerly V-REP): a Versatile and Scalable Robot
  Simulation Framework}.
\newblock In \emph{IEEE International Conference on Intelligent Robots and
  Systems (IROS)}, 2013.

\bibitem[Quattoni and Torralba(2009)]{quattoni2009recognizing}
A.~Quattoni and A.~Torralba.
\newblock Recognizing indoor scenes.
\newblock In \emph{IEEE Conference on Computer Vision and Pattern Recognition
  (CVPR)}. IEEE, 2009.

\bibitem[Finn et~al.(2017)Finn, Yu, Zhang, Abbeel, and Levine]{finn2017one}
C.~Finn, T.~Yu, T.~Zhang, P.~Abbeel, and S.~Levine.
\newblock {One-Shot Visual Imitation Learning via Meta-Learning}.
\newblock In \emph{Conference on Robot Learning (CoRL)}, 2017.

\bibitem[Paszke et~al.(2019)]{NEURIPS2019_9015}
A.~Paszke et~al.
\newblock {PyTorch: An Imperative Style, High-Performance Deep Learning
  Library}.
\newblock In \emph{Conference on Neural Information Processing Systems
  (NeurIPS)}. 2019.

\bibitem[Bradski(2000)]{opencv_library}
G.~Bradski.
\newblock {The OpenCV Library}.
\newblock \emph{Dr. Dobb's Journal of Software Tools}, 2000.

\bibitem[apr()]{april-tag-sim-experiment}
{Pupil-AprilTags}.
\newblock URL \url{https://pypi.org/project/pupil-apriltags/}.

\bibitem[Wang and Olson(2016)]{Wang2016}
J.~Wang and E.~Olson.
\newblock {AprilTag 2: Efficient and robust fiducial detection}.
\newblock In \emph{IEEE International Conference on Intelligent Robots and
  Systems (IROS)}, 2016.

\bibitem[Malyuta(2017)]{malyuta:2017mt}
D.~Malyuta.
\newblock {Guidance, Navigation, Control and Mission Logic for Quadrotor
  Full-cycle Autonomy}.
\newblock Master thesis, Jet Propulsion Laboratory, 2017.

\bibitem[Nair and Hinton(2010)]{nair2010rectified}
V.~Nair and G.~E. Hinton.
\newblock {Rectified Linear Units Improve Restricted Boltzmann Machines}.
\newblock In \emph{International Conference on Machine Learning (ICML)}, 2010.

\bibitem[Ioffe and Szegedy(2015)]{ioffe2015batch}
S.~Ioffe and C.~Szegedy.
\newblock Batch normalization: Accelerating deep network training by reducing
  internal covariate shift.
\newblock In \emph{International Conference on Machine Learning (ICML)}, 2015.

\bibitem[Srivastava et~al.(2014)Srivastava, Hinton, Krizhevsky, Sutskever, and
  Salakhutdinov]{srivastava2014dropout}
N.~Srivastava, G.~Hinton, A.~Krizhevsky, I.~Sutskever, and R.~Salakhutdinov.
\newblock Dropout: a simple way to prevent neural networks from overfitting.
\newblock \emph{The journal of Machine Learning Research (JMLR)}, 2014.

\bibitem[Kingma and Ba(2014)]{kingma2014adam}
D.~P. Kingma and J.~Ba.
\newblock Adam: A method for stochastic optimization.
\newblock \emph{arXiv e-prints}, 2014.

\bibitem[Minai and Williams(1993)]{minai1993derivatives}
A.~A. Minai and R.~D. Williams.
\newblock On the derivatives of the sigmoid.
\newblock \emph{Neural Networks}, 1993.

\bibitem[Lin et~al.(2017)Lin, Goyal, Girshick, He, and Dollár]{focal-loss}
T.-Y. Lin, P.~Goyal, R.~Girshick, K.~He, and P.~Dollár.
\newblock {Focal Loss for Dense Object Detection}.
\newblock In \emph{IEEE International Conference on Computer Vision (ICCV)},
  2017.

\end{thebibliography}
\clearpage

\begin{appendices}

\section{Training Details}

For all our models, we attempt to keep the architectures used as close as reasonably possible to each other to allow for a fair comparison. We use three core components to create all our architectures: (1) an encoder network  using RGB images to output some lower-dimensional abstract feature map, (2) a decoder network that upsamples these abstract feature maps back to the original resolution, and (3) a Multi-Layer Perceptron (MLP)  with dense layers processing these abstract feature maps. As such, the main differences in our architectures occur in (1) using dense layers or convolutional decoders to process the abstract feature maps obtained from the encoders, and (2) the number of output channels in the convolutional decoders which is model dependent.

\begin{figure}
    \centering
    \includegraphics[width=\linewidth]{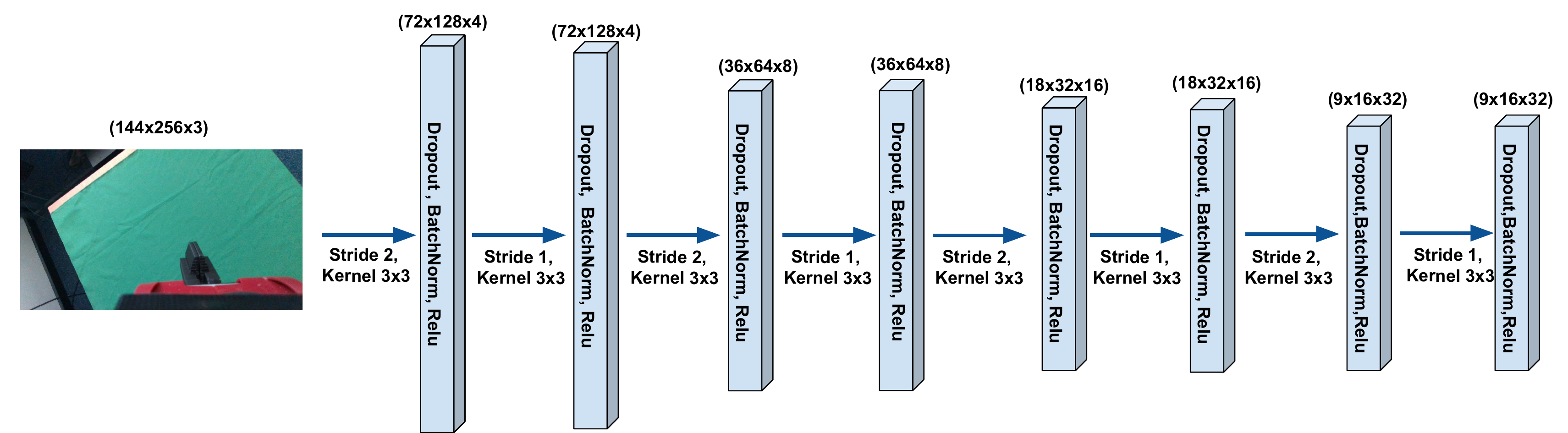}
    \caption{Illustration of the encoder architecture.}
    \label{fig:encoder}
   
\end{figure}

Our encoder network depicted in fig.~\ref{fig:encoder} and described in table~\ref{tab:encoder}. It takes as input a $144 \times 256$ RGB image and has a series of convolutional layers with ReLU~\cite{nair2010rectified} activations, batchnorm~\citep{ioffe2015batch}, and dropout~\citep{srivastava2014dropout} layers. The dropout probability is set to 0.25, and the convolutional layers use a kernel size of $(3\times3)$. We did not include a bias component in our layers, as we found this to be better early in our experimentation. The stride of the convolution alternates between 2 and 1. Padding is set so that a stride of 2 will half the input feature map's resolution, while 1 will keep the resolution the same.

\begin{figure}
    \centering
    \includegraphics[width=\linewidth]{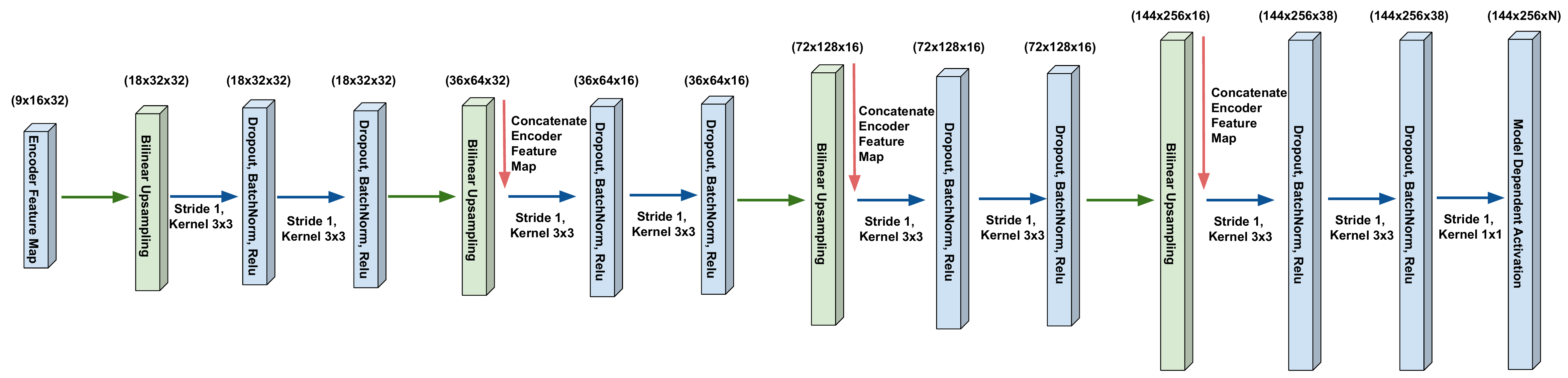}
    \caption{Illustration of the decoder architecture.}
    \label{fig:decoder}
\end{figure}

Our decoder network is depicted in fig.~\ref{fig:decoder} and described in table~\ref{tab:decoder}. It uses a series of non-parametric, bilinear upsampling layers followed by convolutional layers to increase the resolution of the abstract feature map that is the output of our encoder to the original resolution. Each bilinear upsampling layer doubles the spatial resolution of the input feature map, and its output is concatenated with the corresponding features from the encoder in a U-Net-like fashion~\citep{unet}. All convolution layers apart from the last one have ReLU~\cite{nair2010rectified} activations, batchnorm~\citep{ioffe2015batch}, dropout of $0.25$~\citep{srivastava2014dropout}, $(3\times3)$ kernels , stride 1, and padding set to keep the spatial resolution unchanged. Similarly to the encoder network, we did not include a bias component in our layers. The final convolution layer has a kernel size of $(1\times1)$, no batchnorm or dropout, and its activation and spatial resolution are model-dependent (see below).

Our fully connected MLP is depicted in fig.~\ref{fig:mlp} and described in table~\ref{tab:mlp} . It has three hidden fully connected layers of 16 neurons each, with batchnorm~\citep{ioffe2015batch}, $0.25$ dropout~\citep{srivastava2014dropout}, and ReLU~\cite{nair2010rectified} activations. The output layer of the network is a simple linear layer, with no batchnorm or dropout, and directly predicts the translation and orientation encoding of the extrinsic matrix. 

For training all our networks, we use the Adam~\citep{kingma2014adam} optimiser, a batch size of $64$, and a learning rate of $10^{-4}$. We also use a learning rate scheduler that reduces the learning rate by a factor of $0.75$ if performance stagnates. Finally, all RGB images are mapped to the range $[0,1]$ and then normalised before being propagated through the networks.

\begin{table}[b]
\hspace{-1cm}
\footnotesize
\centering
\caption{Table detailing our encoder architecture.}\label{tab:encoder}
\begin{tabular}{ccccc}
\multicolumn{1}{c|}{\textbf{Block}} & \multicolumn{1}{c|}{\textbf{Layers}}            & \multicolumn{1}{c|}{\textbf{Parameters}}                & \multicolumn{1}{c|}{\textbf{Output Size}} & \textbf{Activation} \\ \hline
\multicolumn{1}{c|}{1}              & \multicolumn{1}{c|}{Conv2D, Batchnorm, Dropout} & \multicolumn{1}{c|}{Kernel 3x3, Stride 2, Dropout 0.25} & \multicolumn{1}{c|}{72x128x4}                   & ReLU                \\
\multicolumn{1}{c|}{2}              & \multicolumn{1}{c|}{Conv2D, Batchnorm, Dropout} & \multicolumn{1}{c|}{Kernel 3x3, Stride 1, Dropout 0.25} & \multicolumn{1}{c|}{72x128x4}                   & ReLU                \\
\multicolumn{1}{c|}{3}              & \multicolumn{1}{c|}{Conv2D, Batchnorm, Dropout} & \multicolumn{1}{c|}{Kernel 3x3, Stride 2, Dropout 0.25} & \multicolumn{1}{c|}{36x64x8}                    & ReLU                \\
\multicolumn{1}{c|}{4}              & \multicolumn{1}{c|}{Conv2D, Batchnorm, Dropout} & \multicolumn{1}{c|}{Kernel 3x3, Stride 1, Dropout 0.25} & \multicolumn{1}{c|}{36x64x8}                    & ReLU                \\
\multicolumn{1}{c|}{5}              & \multicolumn{1}{c|}{Conv2D, Batchnorm, Dropout} & \multicolumn{1}{c|}{Kernel 3x3, Stride 2, Dropout 0.25} & \multicolumn{1}{c|}{18x32x16}                   & ReLU                \\
\multicolumn{1}{c|}{6}              & \multicolumn{1}{c|}{Conv2D, Batchnorm, Dropout} & \multicolumn{1}{c|}{Kernel 3x3, Stride 1, Dropout 0.25} & \multicolumn{1}{c|}{18x32x16}                   & ReLU                \\
\multicolumn{1}{c|}{7}              & \multicolumn{1}{c|}{Conv2D, Batchnorm, Dropout} & \multicolumn{1}{c|}{Kernel 3x3, Stride 2, Dropout 0.25} & \multicolumn{1}{c|}{9x16x32}                    & ReLU                \\
\multicolumn{1}{c|}{8}              & \multicolumn{1}{c|}{Conv2D, Batchnorm, Dropout} & \multicolumn{1}{c|}{Kernel 3x3, Stride 1, Dropout 0.25} & \multicolumn{1}{c|}{9x16x32}                    & ReLU                
\end{tabular}

\end{table}

\begin{table}
\footnotesize
\centering
\caption{Table detailing our decoder architecture.}\label{tab:decoder}
\begin{tabular}{lllll}
\multicolumn{1}{c|}{\textbf{Block}} & \multicolumn{1}{c|}{\textbf{Layers}}            & \multicolumn{1}{c|}{\textbf{Parameters / Other}}                                                                       & \multicolumn{1}{c|}{\textbf{Output Size}} & \textbf{Activation} \\ \hline
\multicolumn{1}{c|}{1}              & \multicolumn{1}{c|}{Bilinear Upsampling}        & \multicolumn{1}{c|}{}                                                                                                  & \multicolumn{1}{c|}{18x32x32}                   &                     \\
\multicolumn{1}{c|}{2}              & \multicolumn{1}{c|}{Conv2D, Batchnorm, Dropout} & \multicolumn{1}{c|}{Kernel 3x3, Stride 1, Dropout 0.25}                                                                & \multicolumn{1}{c|}{18x32x32}                   & ReLU                \\
\multicolumn{1}{c|}{3}              & \multicolumn{1}{c|}{Conv2D, Batchnorm, Dropout} & \multicolumn{1}{c|}{Kernel 3x3, Stride 1, Dropout 0.25}                                                                & \multicolumn{1}{c|}{18x32x32}                   & ReLU                \\
\multicolumn{1}{c|}{4}              & \multicolumn{1}{c|}{Bilinear Upsampling}        & \multicolumn{1}{c|}{\begin{tabular}[c]{@{}l@{}}Skip Connection Concatenation\\ with Encoder Feature Maps\end{tabular}} & \multicolumn{1}{c|}{36x64x32}                   &                     \\
\multicolumn{1}{c|}{5}              & \multicolumn{1}{c|}{Conv2D, Batchnorm, Dropout} & \multicolumn{1}{c|}{Kernel 3x3, Stride 1, Dropout 0.25}                                                                & \multicolumn{1}{c|}{36x64x32}                   & ReLU                \\
\multicolumn{1}{c|}{6}              & \multicolumn{1}{c|}{Conv2D, Batchnorm, Dropout} & \multicolumn{1}{c|}{Kernel 3x3, Stride 1, Dropout 0.25}                                                                & \multicolumn{1}{c|}{36x64x32}                   & ReLU                \\
\multicolumn{1}{c|}{7}              & \multicolumn{1}{c|}{Bilinear Upsampling}        & \multicolumn{1}{c|}{\begin{tabular}[c]{@{}l@{}}Skip Connection Concatenation\\ with Encoder Feature Maps\end{tabular}} & \multicolumn{1}{c|}{72x128x32}                  &                     \\
\multicolumn{1}{c|}{8}              & \multicolumn{1}{c|}{Conv2D, Batchnorm, Dropout} & \multicolumn{1}{c|}{Kernel 3x3, Stride 1, Dropout 0.25}                                                                & \multicolumn{1}{c|}{72x128x24}                  & ReLU                \\
\multicolumn{1}{c|}{9}              & \multicolumn{1}{c|}{Conv2D, Batchnorm, Dropout} & \multicolumn{1}{c|}{Kernel 3x3, Stride 1, Dropout 0.25}                                                                & \multicolumn{1}{c|}{72x128x24}                  & ReLU                \\
\multicolumn{1}{c|}{10}             & \multicolumn{1}{c|}{Bilinear Upsampling}        & \multicolumn{1}{c|}{\begin{tabular}[c]{@{}l@{}}Skip Connection Concatenation\\ with Encoder Feature Maps\end{tabular}} & \multicolumn{1}{c|}{144x256x24}                 &                     \\
\multicolumn{1}{c|}{11}             & \multicolumn{1}{c|}{Conv2D, Batchnorm, Dropout} & \multicolumn{1}{c|}{Kernel 3x3, Stride 1, Dropout 0.25}                                                                & \multicolumn{1}{c|}{144x256x42}                 & ReLU                \\
\multicolumn{1}{c|}{12}             & \multicolumn{1}{c|}{Conv2D, Batchnorm, Dropout} & \multicolumn{1}{c|}{Kernel 3x3, Stride 1, Dropout 0.25}                                                                & \multicolumn{1}{c|}{144x256x42}                 & ReLU                \\
\multicolumn{1}{c|}{13}             & \multicolumn{1}{c|}{Conv2D}                     & \multicolumn{1}{c|}{Kernel 1x1, Stride 1}                                                                              & \multicolumn{1}{c|}{}                           &                     
\end{tabular}
\end{table}

\begin{table}
\small
\centering
\caption{Table detailing the MLP architecture.}\label{tab:mlp}
\begin{tabular}{lllll}
\multicolumn{1}{c|}{\textbf{Block}} & \multicolumn{1}{c|}{\textbf{Layers}}            & \multicolumn{1}{c|}{\textbf{Parameters / Other}} & \multicolumn{1}{c|}{\textbf{Output Size}} & \textbf{Activation} \\ \hline
\multicolumn{1}{c|}{1}              & \multicolumn{1}{c|}{Linear, Batchnorm, Dropout} & \multicolumn{1}{c|}{Dropout 0.25}                & \multicolumn{1}{c|}{16}                         & ReLU                \\
\multicolumn{1}{c|}{2}              & \multicolumn{1}{c|}{Linear, Batchnorm, Dropout} & \multicolumn{1}{c|}{Dropout 0.25}                & \multicolumn{1}{c|}{16}                         & ReLU                \\
\multicolumn{1}{c|}{3}              & \multicolumn{1}{c|}{Linear, Batchnorm, Dropout} & \multicolumn{1}{c|}{Dropout 0.25}                & \multicolumn{1}{c|}{16}                         & ReLU                \\
\multicolumn{1}{c|}{4}              & \multicolumn{1}{c|}{Linear}                     & \multicolumn{1}{c|}{}                            & \multicolumn{1}{c|}{9}                          &                     
\end{tabular}
\end{table}

\begin{figure}
    \centering
    \includegraphics[width=\linewidth]{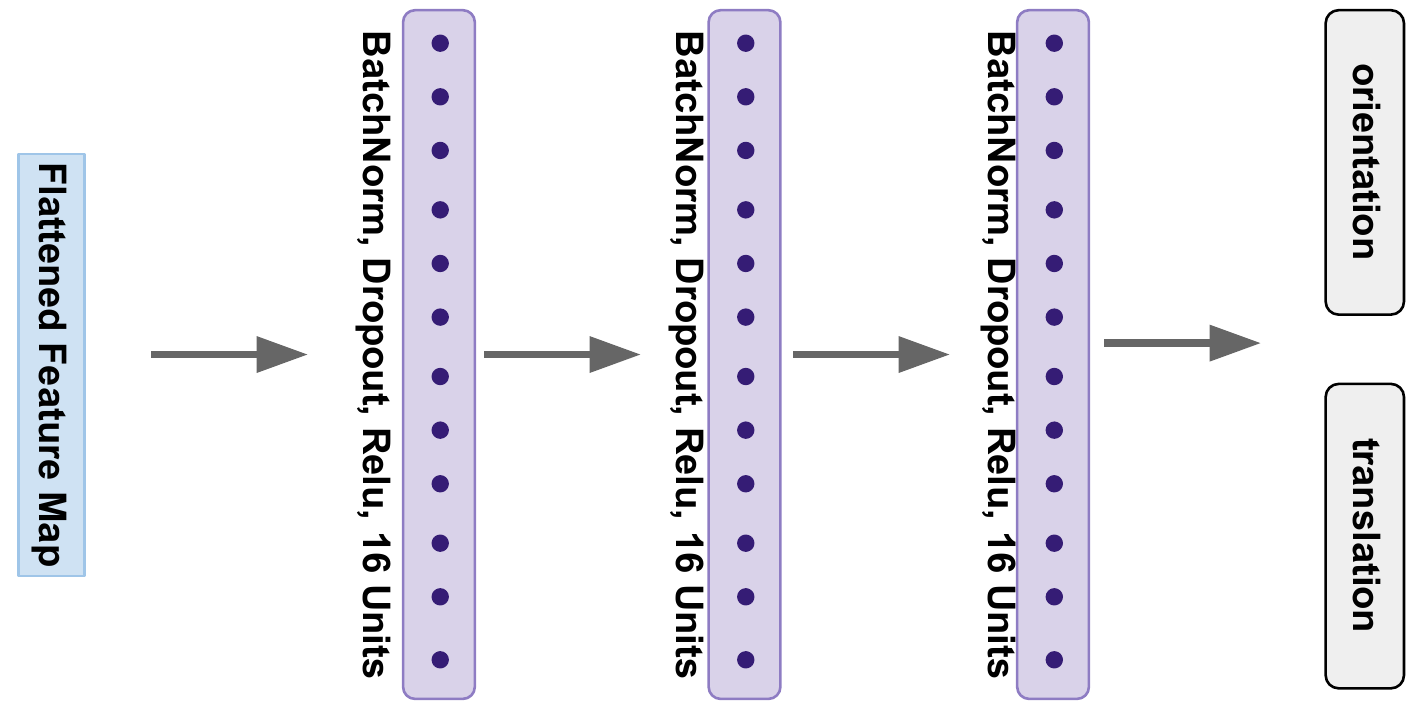}
    \caption{Illustration of the MLP architecture.}
    \label{fig:mlp}
\end{figure}
\subsection{Direct Regression Details}
For our direct regression model, we use our encoder network followed by our MLP network. The loss used was the mean squared error (MSE) between the network outputs and the ground truth extrinsic matrix encodings obtained from the simulated data.  Finally, each output dimension was independently normalised to the range $[-1,1]$, using min-max normalisation and the training dataset statistics. 
\subsection{Sparse Correspondence Model Details}
Our sparse correspondence model consists of our encoder, followed by our decoder with its number of output channels being set to the number of keypoints used, and its output activation set to the spatial-soft-argmax activation~\cite{finn2016deep}.  The loss used was an equally-weighted linear combination of the L1 loss and the MSE loss between the network outputs and the ground truth keypoint locations obtained from the simulated data. Finally, all the keypoint locations are normalised in the range $[-1,1]$  using the image dimensions, with any keypoint coordinate that appears out of frame (hence outside of the $[-1,1]$ range) projected back to $[-1,1]$. Before using the predicted 2D keypoint location to solve for the camera pose with PnP, we reject all keypoints less than 1\% away from the image rim. 

\subsection{Dense Correspondence Model Details}
For our dense correspondence model, the ICP initialisation network is the same as the direct regression network. The segmentation network consists of our encoder followed by our decoder, with a single output channel with the sigmoid activation~\citep{minai1993derivatives}. The loss we use for this segmentation network is the $\alpha-$balanced variant of the Focal Loss~\citep{focal-loss}, with the focusing parameter $\gamma=2$, and with the weighting factor $\alpha$  proportional to the inverse class frequency.  Finally, the depth regression network consists of our encoder network followed by our decoder network, with a single output channel fitted with a ReLU~\cite{nair2010rectified} activation function. The loss used for the depth regression is the L1 loss between the predicted depth map and the ground truth depth obtained from the simulated data, that is calculated over the union between the predicted and ground truth segmentation mask. Finally, the depth maps are normalised in the $[0,1]$ range, using min-max normalisation and the training dataset statistics. 
\section{Dataset Generation Details}\label{sec:dataset-generation-details}
\begin{figure}
    \centering
    \includegraphics[width=\linewidth]{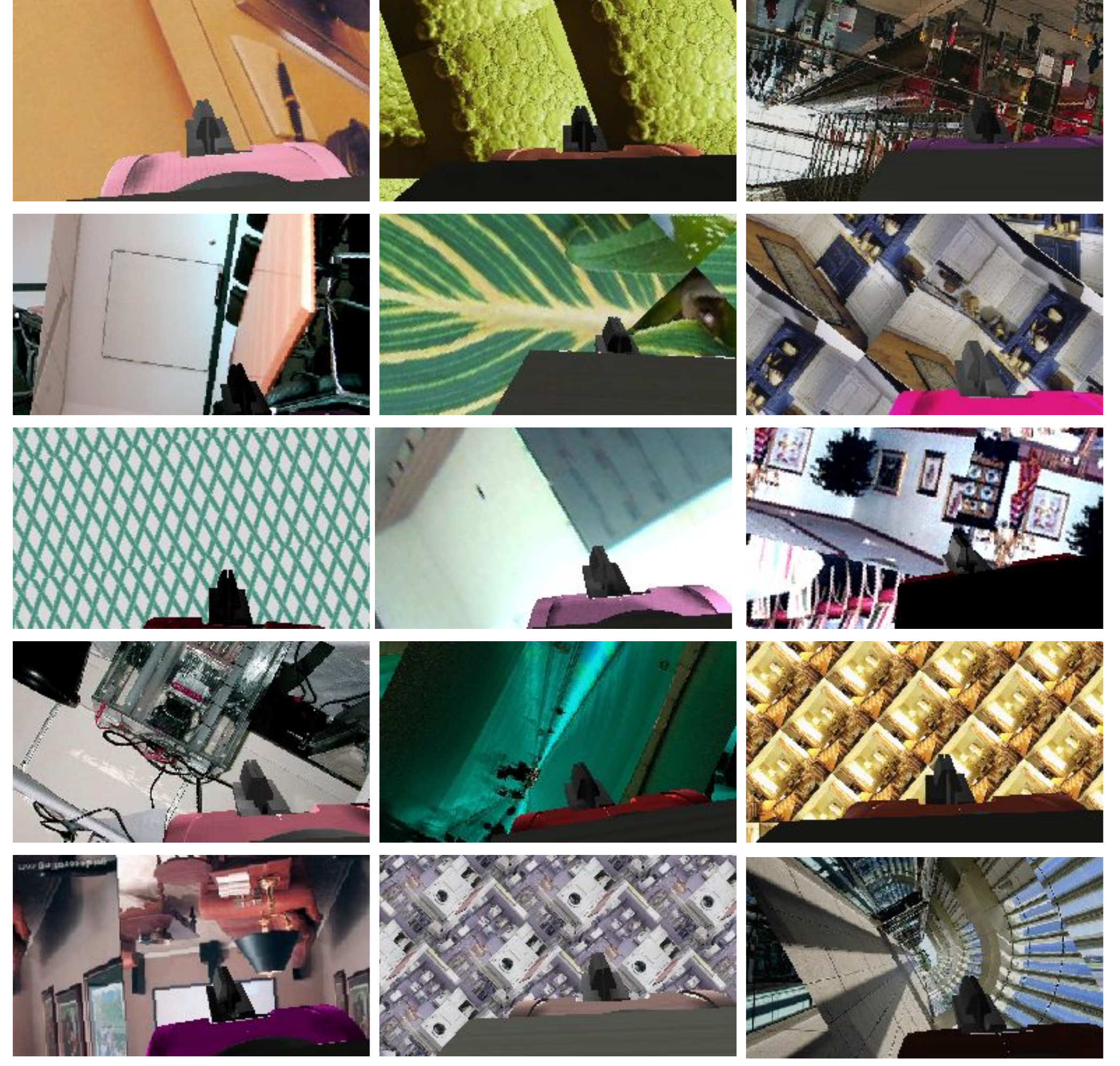}
    \caption{Examples of Domain Randomised Images}
    \label{fig:dr_examples}
    \vspace{-0.5cm}
\end{figure}

\footnotetext[2]{{Textures modified from \url{https://github.com/tianheyu927/mil/blob/master/scripts/get_data.sh}, made available by~\cite{finn2017one}.}}

All our datasets consist of $10~000$ images generated in simulation, examples of which can be seen in fig.~\ref{fig:dr_examples}. For each image in the dataset, we randomise the position of the camera in a $3cm\times3cm\times3cm$ volume and its orientation by $\pm 5^{\circ}$ around each axis. The mean of the randomisation range is estimated using the in-built Sawyer camera (which is already defined in the URDF) by taking images of the same AprilTag~\citep{Wang2016,malyuta:2017mt} from both the in-built camera and our wrist-mounted camera and calculating the transformation between the two. We then apply random grayscale textures\footnotemark[2] to each of the components of the end-effector in simulation, and randomise their colours by (1) changing the hue as $h = h_{original}*(1.0+ e_h)$, (2) changing the brightness as $b = b_{original}*(1.0+ e_b)$, and (3) changing the saturation as $s =s_{original}*(1.0+ e_s)$, where $e_h, e_b$, and $e_s$ are sampled uniformly from the ranges $[-0.2, 0.2]$, $[-0.45, 0.45]$ and  $[-0.6, 0.5]$, respectively. We also randomise the position of two point light sources, sampled uniformly in a spherical shell around the gripper, with a radius varying in $[1.5m, 4.0 m]$. Our simulator generates images at a $480\times848$ resolution, similarly to our real-world sensor. Before applying the background randomisation, we downsample these images to $144\times256$, which is our neural network input resolution. When applying the background images for background randomisation, we randomly rotate them in the range $[0^{\circ},360^{\circ}]$, and tile them when needed to avoid portions of the image with no background applied. Finally, we apply a colour jitter~\citep{NEURIPS2019_9015} operation to the full image with brighness variation of 0.2, contrast variation of 0.2, saturation variation of 0.2 and hue variation of 0.05.

\section{Fusing Multiple Estimates}

\begin{algorithm}
\SetAlgoLined
\DontPrintSemicolon
\caption{Fusing multiple estimates}\label{alg:fusing-mesurements}
\KwIn{Individual estimates $\{\Tilde{T}_{EC}^i=[\Tilde{R}_{EC}^i | \Tilde{t}_{EC}^i]\}_{i=1}^N$}
\KwOut{A fused estimate $\Bar{T}_{EC}$}
\nl Initialise $\xi =[]$\;
\For{$k=1,..., N$}{
    \nl $\xi \leftarrow \xi \cup \left(\Tilde{t}_{EC}^k, \phi^k \right)$\ where $\phi_k$ are the intrinsic Euler XYZ angles that represent the rotation matrix $\Tilde{R}_{EC}^k$;
    }
\nl Calculate the mean $\mu = \text{mean}(\xi) \in \mathds{R}^6$ and estimate the covariance $\Sigma = \text{Cov}(\xi) \in \mathds{R}^{6\times 6}$ \;
\nl Discard 20\% of estimates in $\xi$ with the lowest probability density under the PDF $\mathcal{N}(\mu, \Sigma)$ \;
\nl Calculate the mean of the remaining estimates $\Bar{\xi} = \text{mean}(\xi)=(\Bar{t}, \Bar{Euler})$ \;
\nl $\Bar{T}_{EC}=[\text{Euler2Rot}(\Bar{Euler})| \Bar{t}]$ \;
\end{algorithm}
Algorithm \ref{alg:fusing-mesurements} describes the procedure used to fuse multiple camera pose estimates into a single estimate. This algorithm is initialised with a list of camera pose estimates, $\{\Tilde{T}_{EC}^i=[\Tilde{R}_{EC}^i | \Tilde{t}_{EC}^i]\}_{i=1}^N$, and creates an empty list $\xi=[]$. It then iterates through all individual estimates $\Tilde{T}_{EC}^k$ for $k=1,..., N$, and represents them as 6D vectors $ \left(\Tilde{t}_{EC}^k, \phi_k\right)$, where $\phi_k$ are the intrinsic Euler XYZ angles that represent the rotation matrix $\Tilde{R}_{EC}^k$, and stores these representations in the list $\xi$. After mapping all of the input estimates to a lower dimensional representation, line 3 in the pseudocode computes the mean and an unbiased estimate of the covariance matrix of all of the estimates. In line 4, we then assume that the data follows a Gaussian distribution and discard $20\%$ of the estimates with the lowest probability density under this Gaussian model in order to remove potential outliers. In line 5, we then calculate the mean camera pose in the 6D space, and in line 6 we transform this lower dimensional representation of the mean estimate back to a $4\times4$ homogeneous transformation matrix.
%
%
\section{Experiments}

We set up a single simulation environment for all our learning-based methods, and generate a single dataset to train them. Dataset generation took approximately 30 minutes given the ability to parallelise data collection. Training the direct regression model took approximately 30 minutes, the sparse correspondence model approximately 3.5 hours, and the dense correspondence model approximately 3 hours. Finally, we performed early stopping with a 5\% validation subset.

In the following sections we use the following method abbreviations: \textbf{D}irect \textbf{R}egression (DR), \textbf{S}parse \textbf{C}orrespondense method (SC), and \textbf{D}ense \textbf{C}orrespondence method (DC). When $(fusion)$ is specified, we indicate that our fusion method for aggregating multiple estimates was used.
%
%
%
\subsection{Simulated Experiment}
%
%
\begin{algorithm}
\SetAlgoLined
\DontPrintSemicolon
\caption{Simulated Experiment}\label{alg:simulated-experiment}
\nl $\epsilon_{pos}^{\textbf{method}}=[]$ and $\epsilon_{ori}^{\textbf{method}}=[]$ for each \textbf{method} $\in$ [Tsai~\citep{tsai1989new}, Park~\citep{park1994robot}, Horaud~\citep{horaud1995hand}, Daniilidis~\citep{daniilidis1999hand}, DR, DR (fusion), SC, SC (fusion), DC, DC (fusion)] \;
\nl \For{$i=1,..., N$}{
    \nl Sample and set a camera to end-effector pose $T_{EC}=[R_{EC}|t_{EC}]$ \;
    \nl Collect a dataset $\{I_j, T_{BE}^j, \Tilde{T}_{CO}^j\}_{j=1}^{15}$ of 15 RGB images $I$, end-effector to robot base poses $T_{BE}$, and estimated calibration object to camera poses $\Tilde{T}_{CO}$\;
    \nl \For{\textbf{method} $\in$ [Tsai~\citep{tsai1989new}, Park~\citep{park1994robot}, Horaud~\citep{horaud1995hand}, Daniilidis~\citep{daniilidis1999hand}]}{
        \nl $\Tilde{T}_{EC} = [\Tilde{R}_{EC}|\Tilde{t}_{EC}] \leftarrow \text{use\_}\textbf{method}\text{\_for\_calibration}\left(\{T_{BE}^j, \Tilde{T}_{CO}^j\}_{j=1}^{15}\right)$ \;
        \nl $\epsilon_{pos}^{\textbf{method}} \leftarrow \epsilon_{pos}^{\textbf{method}} \cup e_{t}(\Tilde{t}_{EC}, t_{EC})$ \;
        \nl $\epsilon_{ori}^{\textbf{method}} \leftarrow \epsilon_{ori}^{\textbf{method}} \cup e_{R}(\Tilde{R}_{EC}, R_{EC})$ \;
        }
    \nl \For{\textbf{method} $\in$ [DR, SC, DC]}{
        \nl $T=[]$ \;
        \nl \For{$k=1, ..., 15$}{
            \nl $\Tilde{T}_{EC}^k=[\Tilde{R}_{EC}|\Tilde{t}_{EC}]\leftarrow \text{use\_}\textbf{method}\text{\_for\_calibration}(I_k)$ \;
            \nl $T \leftarrow T \cup \Tilde{T}_{EC}^k$ \;
            \nl $\epsilon_{pos}^{\textbf{method}} \leftarrow \epsilon_{pos}^{\textbf{method}} \cup e_{t}(\Tilde{t}_{EC}, t_{EC})$ \;
            \nl $\epsilon_{ori}^{\textbf{method}} \leftarrow \epsilon_{ori}^{\textbf{method}} \cup e_{R}(\Tilde{R}_{EC}, R_{EC})$ \;  
            }
    \nl $\Bar{T}_{EC} = [\Bar{R}_{EC} | \Bar{t}_{EC}] \leftarrow \text{fusion}(T)$ \;
    \nl $\epsilon_{pos}^{\text{\textbf{method} (fusion)}} \leftarrow \epsilon_{pos}^{\text{\textbf{method} (fusion)}} \cup e_{t}(\Bar{t}_{EC}, t_{EC})$ \;
    \nl $\epsilon_{ori}^{\text{\textbf{method} (fusion)}} \leftarrow \epsilon_{ori}^{\text{\textbf{method} (fusion)}} \cup e_{R}(\Bar{R}_{EC}, R_{EC})$ \;   
    }
    }
\nl $\varepsilon_t^{\textbf{method}} \leftarrow \text{mean}(\epsilon_{pos}^{\textbf{method}})$ for all methods \;
\nl $\varepsilon_R^{\textbf{method}} \leftarrow \text{mean}(\epsilon_{ori}^{\textbf{method}})$ for all methods \;
\end{algorithm}
Algorithm \ref{alg:simulated-experiment} outlines the procedure used to evaluate all methods in simulation. In line 1, the algorithm initialises an empty list to store the position and orientation error for each of the methods independently, so that these errors can later be averaged across $N$ different extrinsic matrices. The position error is defined as
\begin{equation*}
    e_{t}(\Tilde{t}, t) = ||\Tilde{t} - t||_2
\end{equation*}
where $||\cdot||_2$ is the L2 norm. The orientation error is defined as
\begin{equation*}
    e_{R}(\Tilde{R}, R)=\theta
\end{equation*}
where $\theta$ is the angle of rotation from the axis-angle representation of the rotation matrix $R_{\Delta}$ that satisfies the relationship $R=R_{\Delta}\Tilde{R}$.

Once the lists are initialised, the outer for loop that extends from line 2 to line 18 of the algorithm is responsible for collecting position and orientation errors for various samples of the camera to end-effector pose. Within this for loop, in line 3, a camera to end-effector pose is sampled and set. Then, in line 4, a dataset of 15 corresponding RGB images and end-effector to robot base poses is collected by moving the end-effector around the AprilTag. The corresponding AprilTag to camera poses are estimated from the RGB images with the AprilTags3 library~\citep{april-tag-sim-experiment,Wang2016,malyuta:2017mt}.

Between lines 5 and 8, the algorithm then estimates the camera to end-effector pose using each of the classical methods and evaluates each of the estimates. Then, between lines 9 and 18, the algorithm evaluates all of the deep learning-based methods. For each method, the algorithm begins by creating an empty list (line 10) that is used to store all individual estimates. It then iterates through all of the individual images and evaluates all of the single image estimates (lines 11-15). It then fuses all of the individual estimates using algorithm \ref{alg:fusing-mesurements} and evaluates the fused estimate (lines 16-18). The algorithm ends by computing the mean of the position and orientation errors for each of the methods (lines 19 and 20).

We stress that even though the AprilTag is visible in all images used to estimate the camera to end-effector pose, it is not used by the learned methods. The same images are used to evaluate the learned methods as the ones to estimate the tag to camera pose that is required by the classical methods only to ensure a fair evaluation. 
%
%
\subsection{Real World Experiment}
%
%
%
\begin{algorithm}
\SetAlgoLined
\DontPrintSemicolon
\caption{Real World Experiment}\label{alg:real-world-experiment}
\nl $T^{\textbf{method}}=[]$ for \textbf{method} $\in$ [Tsai~\citep{tsai1989new}, Park~\citep{park1994robot}, Horaud~\citep{horaud1995hand}, Daniilidis~\citep{daniilidis1999hand}, DR, DR (fusion), SC, SC (fusion), DC, DC (fusion)] \;
\nl Collect a training data bank $\mathcal{D}_{train}=\{I_i, T_{BE}^i, \Tilde{T}_{CO}^i\}_{i=1}^{40}$ \;
\nl Collect an evaluation dataset $\mathcal{D}_{eval}=\{ T_{BE}^i, \Tilde{T}_{CO}^i\}_{i=1}^{60}$ \;
\nl \For{$i=1, ..., 40$}{
    \nl Sample a dataset $\mathcal{D}=\{I_j, T_{BE}^j, \Tilde{T}_{CO}^j\}_{j=1}^{15}\in \mathcal{D}_{train}$ \;
    \nl \For{\textbf{method} $\in$ [Tsai~\citep{tsai1989new}, Park~\citep{park1994robot}, Horaud~\citep{horaud1995hand}, Daniilidis~\citep{daniilidis1999hand}]}{
        \nl $\Tilde{T}_{EC}^i \leftarrow \text{use\_}\textbf{method}\text{\_for\_calibration}\left(\{T_{BE}^j, \Tilde{T}_{CO}^j\}_{j=1}^{15}\right)$ \;
        \nl $T^{\textbf{method}} \leftarrow T^{\textbf{method}} \cup \Tilde{T}_{EC}^i$
        }
    \nl \For{\textbf{method} $\in$ [DR, SC, DC]}{
        \nl $\xi = []$ \;
        \nl \For{$k=1, ..., 15$}{
            \nl $\Tilde{T}_{EC}^k= \leftarrow \text{use\_}\textbf{method}\text{\_for\_calibration}(I_k)$ \;
            \nl $\xi \leftarrow \xi \cup \Tilde{T}_{EC}^k$ \;
            }
        \nl $\Bar{T}_{EC} = \leftarrow \text{fusion}(\xi)$ \;
        \nl $T^{\text{\textbf{method} (fusion)}} \leftarrow T^{\text{\textbf{method} (fusion)}} \cup \Bar{T}_{EC}$
        }
    }
\nl \For{$I_i\in \mathcal{D}_{train}$}{
    \nl \For{\textbf{method} $\in$ [DR, SC, DC]}{
        \nl $\Tilde{T}_{EC}^i= \leftarrow \text{use\_}\textbf{method}\text{\_for\_calibration}(I_i)$ \;
        \nl $T^{\textbf{method}} \leftarrow T^{\textbf{method}} \cup \Tilde{T}_{EC}^i$ \;
        }
    }
\nl \For{\textbf{method} $\in$ [Tsai~\citep{tsai1989new}, Park~\citep{park1994robot}, Horaud~\citep{horaud1995hand}, Daniilidis~\citep{daniilidis1999hand}, DR, DR (fusion), SC, SC (fusion), DC, DC (fusion)]}{
    $t=[]$ \;
    \nl \For{$\Tilde{T}_{EC}^k \in T^{\textbf{method}}$}{
        \nl \For{$\{T_{BE}^i, \Tilde{T}_{CO}^i\} \in \mathcal{D}_{eval}$}{
            $\Tilde{T}_{BO}^{ik} = [\Tilde{R}_{BO}^{ik} | \Tilde{t}_{BO}^{ik}] = T_{BE}^i \Tilde{T}_{EC}^k \Tilde{T}_{CO}^i$ \;
            \nl $t \leftarrow t \cup \Tilde{t}_{BO}^{ik}$ \;

            }
        }
    \nl $\epsilon_{std}^{\textbf{method}} = \text{standard devation}(t)$
    }
\end{algorithm}\textbf{}

The real-world experimental procedure is shown in algorithm \ref{alg:real-world-experiment}. This procedure consists of 4 independent stages. In the first stage (lines 1-3), a data bank and an evaluation dataset of RGB images, end-effector to robot base poses, and estimated AprilTag to camera poses are collected by automatically moving the robot's end-effector around an AprilTag. We estimate the AprilTag poses using the AprilTags3 library~\citep{april-tag-sim-experiment, Wang2016,malyuta:2017mt}. The lengths of the training data bank and evaluation dataset are 40 and 60 respectively. 

In the second stage (lines 4-15), we obtain estimates of the camera to end-effector pose from all methods that require more than a single data point. During this stage, we sample 40 datasets of 15 data points each from the data bank. For each dataset, we estimate the camera to end-effector pose using the classical baselines. We also estimate the camera to end-effector pose from each of the RGB images using each of the learned methods and fuse all 15 estimates into a single estimate using algorithm \ref{alg:fusing-mesurements}. In the third stage (lines 16-19), we obtain the estimates from the learning-based methods for all of the images in the training data bank.

The final stage (lines 20-24) evaluates each of the methods. As in the real world we do not have access to the ground truth camera to end-effector pose, we use an indirect error metric~\cite{tsai1989new, daniilidis1999hand}. Let $T_{BE}$ be the end-effector to robot base pose, $\Tilde{T}_{EC}$ be an estimate of the camera to end-effector pose and $\Tilde{T}_{CO}$ be an estimate of the AprilTag to camera pose. An estimate of the AprilTag to robot base pose can be computed as
\begin{equation*}
    \Tilde{T}_{BO} = T_{BE} \Tilde{T}_{EC} \Tilde{T}_{CO}
\end{equation*}
Now, for a single estimate of the camera to end-effector pose, $\Tilde{T}_{EC}^k$ and for each pair of end-effector to robot base and AprilTag to camera poses, $\{T_{BE}^i, \Tilde{T}_{CO}^i\} \in \mathcal{D}_{eval}$, a different estimate of the AprilTag to robot base pose can be computed, $\Tilde{T}_{BO}^{ik}=[\Tilde{R}_{BO}^{ik} | \Tilde{t}_{BO}^{ik}]$. We use the standard deviation of all estimated AprilTag to robot base positions for a single method, $\Tilde{t}_{BO}^{ik}$ for all $i, k$, as an indirect measure of the accuracy of that method. The higher the precision of the estimated camera to end-effector matrix, the lower the spread of the estimated AprilTag positions in the robot base frame, and hence, the lower the standard deviation.
%
%

\section{Additional Gripper Experiments}

\begin{figure}
    \centering
    \includegraphics[width=\linewidth]{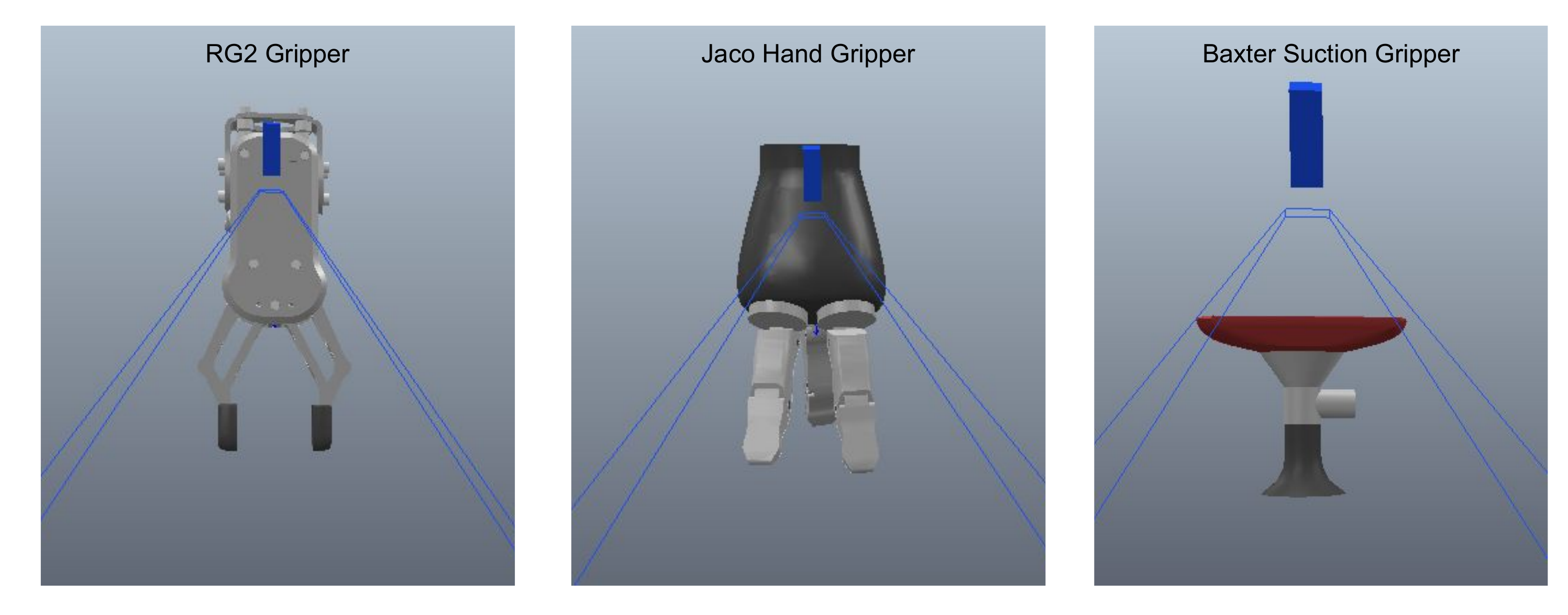}
    \caption{Illustration of additional grippers tested.}
    \label{fig:gips}
   
\end{figure}

In order to ensure there is nothing particular about the Sawyer two-finger gripper we used in our experiments, we further test our best performing method with three additional simulated grippers, which are illustrated in fig.~\ref{fig:gips}. As such, we train the direct regression model on a second two-finger gripper, a three-finger gripper, and a suction gripper. We evaluate the performance on each using our simulation testing procedure and report the results in table~\ref{tab:grips}. We also append the performance on the Sawyer two-finger gripper we reported in the main paper for reference. From the table, see that the test performance on the tree grippers is very similar, which indicates that our method could be applied to any gripper requirement. We note that setting up each new gripper training is very simple and only requires the gripper model to be replaced in simulation, which can be accomplished in a time of the order of 5 minutes. As such, considering the dataset generation and training with the direct regression method, the additional time required to get the direct regression method working on a new gripper is approximately 2 hours in total.

\begin{table}
\small
    \centering
        \begin{tabular}{|l|c|c|}
        \hline
        Gripper &      $\epsilon_t$ [mm] & $\epsilon_R$ [degrees] \\
        \midrule
        Sawyer                     &     $(13.4 \pm 4.1)$ &        $(4.4 \pm 1.4)$\\
        Sawyer (fusion)             &     $(13.4 \pm 4.1)$ &        $(4.4 \pm 1.4)$ \\
        Jaco Hand        &  $(10.1 \pm 4.0)$ &     $ (3.6 \pm 1.4)$ \\
        Jaco Hand (fusion) &  $ (9.9 \pm 3.8)$ &     $(3.5 \pm 1.3)$ \\
        RG2           &    $(10.3 \pm 3.8)$ &      $(4.5 \pm 1.3)$  \\
        RG2 (fusion)    &    $(10.3 \pm 3.8)$ &      $(4.5 \pm 1.3)$  \\
        Baxter Suction           &    $(11.2 \pm 4.3)$ &      $ (3.8 \pm 1.4)$  \\
        Baxter Suction (fusion)    &    $(11.1 \pm 4.2)$ &      $ (3.8 \pm 1.4)$  \\
        \bottomrule
        \end{tabular}
    \caption{Direct Regression performance of four different grippers in simulation.}
    \vspace{-0.5cm}
    \label{tab:grips}
\end{table}

\end{appendices}

\end{document}